%% file: main.tex
\documentclass[10pt,journal,compsoc,x11names]{IEEEtran}
\usepackage[switch]{lineno}
\usepackage{amsfonts}
\usepackage{amsmath, bm}
\usepackage{pifont} 
\usepackage{xspace}
\usepackage{enumitem}
\usepackage{amssymb}
\usepackage{booktabs}
\usepackage[hyphens]{url}

\usepackage{ragged2e}
\usepackage{hyperref}
\usepackage{multirow}
\usepackage{graphicx}
\usepackage{xcolor}
\usepackage{color}
\usepackage{colortbl}
\usepackage{tablefootnote}
\usepackage{pifont}
\usepackage{makecell}
\usepackage[most]{tcolorbox}
\usepackage{framed}
\usepackage{mdframed}
\usepackage{subfigure}
\usepackage{caption}
\usepackage{longtable}
\usepackage{float}
\usepackage{booktabs}

\newcolumntype{H}{>{\setbox0=\hbox\bgroup}c<{\egroup}@{}}

\newcommand{\ignore}[1]{}

\definecolor{gold}{RGB}{205,133,63}
\definecolor{fGreen}{RGB}{34,139,34}
\definecolor{tOrange}{RGB}{255,215,0}
\definecolor{tBlue}{RGB}{135,206,250}
\definecolor{tPink}{RGB}{255,204,204}
\definecolor{tGreen}{RGB}{205,230,199}
\definecolor{tGold}{RGB}{255,215,0}

\usepackage{cite}
\usepackage[numbers,sort&compress]{natbib}

\usepackage{tcolorbox}
\usepackage{pifont}
\usepackage{longtable}
\usepackage{makecell}
\usepackage{multirow} 
\newcommand{\exa}[1]{{#1}}

\ifCLASSINFOpdf
\else
\fi

\begin{document}
\title{A Survey of Interactive Generative Video}

\author{Jiwen Yu$^*$, Yiran Qin$^*$, Haoxuan Che$^*$, \\Quande Liu$^\dagger$, Xintao Wang, Pengfei Wan, Di Zhang, Kun Gai, Hao Chen, Xihui Liu$^\dagger$
\IEEEcompsocitemizethanks{
\IEEEcompsocthanksitem $^*$ indicates equal contributions. $^\dagger$ indicates corresponding authors. 
\IEEEcompsocthanksitem Jiwen Yu, Yiran Qin, and Xihui Liu are with the The University of Hong Kong, Pok Fu Lam, Hong Kong. 
\IEEEcompsocthanksitem Quande Liu, Xintao Wang, Pengfei Wan, Di Zhang, and Kun Gai are with Kuaishou Technology, Shenzhen, China. 
\IEEEcompsocthanksitem Haoxuan Che and Hao Chen are with The Hong Kong University of Science and Technology (HKUST), Hong Kong. 
}%
}

\markboth{}%
{Shell \MakeLowercase{\textit{et al.}}: Bare Advanced Demo of IEEEtran.cls for IEEE Computer Society Journals}

\IEEEtitleabstractindextext{%
\begin{abstract}
Interactive Generative Video (IGV) has emerged as a crucial technology in response to the growing demand for high-quality, interactive video content across various domains. In this paper, we define IGV as a technology that combines generative capabilities to produce diverse high-quality video content with interactive features that enable user engagement through control signals and responsive feedback. 
We survey the current landscape of IGV applications, focusing on three major domains: 1) gaming, where IGV enables infinite exploration in virtual worlds; 2) embodied AI, where IGV serves as a physics-aware environment synthesizer for training agents in multimodal interaction with dynamically evolving scenes; and 3) autonomous driving, where IGV provides closed-loop simulation capabilities for safety-critical testing and validation.
To guide future development, we propose a comprehensive framework that decomposes an ideal IGV system into five essential modules: Generation, Control, Memory, Dynamics, and Intelligence. Furthermore, we systematically analyze the technical challenges and future directions in realizing each component for an ideal IGV system, such as achieving real-time generation, enabling open-domain control, maintaining long-term coherence, simulating accurate physics, and integrating causal reasoning. We believe that this systematic analysis will facilitate future research and development in the field of IGV, ultimately advancing the technology toward more sophisticated and practical applications.
\end{abstract}

}

\maketitle

\IEEEdisplaynontitleabstractindextext

\IEEEpeerreviewmaketitle

\input{IGV_sections/1_intro}

\input{IGV_sections/2_preliminary}
\input{IGV_sections/3_sys_overview}

\input{IGV_sections/4_app1_game}

\input{IGV_sections/5_app2_embodied}

\input{IGV_sections/6_app3_autodrive}

\input{IGV_sections/7_conclusion}

\ifCLASSOPTIONcaptionsoff
  \newpage
\fi

\bibliographystyle{IEEEtran}
\bibliography{newbib}

\end{document}

%% file: IGV_sections/1_intro.tex
\section{Introduction}\label{sec:introduction}
The demand for high-quality interactive video has grown significantly in recent years across various domains, from digital entertainment to industrial applications. This trend reflects the expanding role of video technology in applications such as simulation, decision-making, and content creation. Meanwhile, video generation technology has witnessed remarkable progress~\cite{sora,cogvideox,kling,veo2,seaweed,moviegen,runway,luma,vidu,wan,hunyuanvideo}, driven by recent advances in generative modeling paradigms, particularly diffusion models~\cite{ddpm,score,flow,rectified} and next-token prediction approaches~\cite{videogpt,videopoet,emu3}. Modern video generation systems can now produce remarkably realistic outputs while offering precise control over the generated content, enabling unprecedented opportunities across different fields.

Motivated by these emerging needs and rapid technological developments, we present a comprehensive survey of interactive generative video technology in this paper. To establish a foundation for our discussion, we first introduce the concept of Interactive Generative Video (IGV), which is characterized by two key aspects. First, it is fundamentally generative, using generative models trained on vast video datasets to produce diverse, high-quality, open-domain video content. Second, it is inherently interactive, enabling online user engagement through control signals and responsive feedback, allowing users to achieve specific tasks or experiences through their interactions.

\begin{figure*}[ht]
  \centering
  \includegraphics[width=0.9\linewidth]{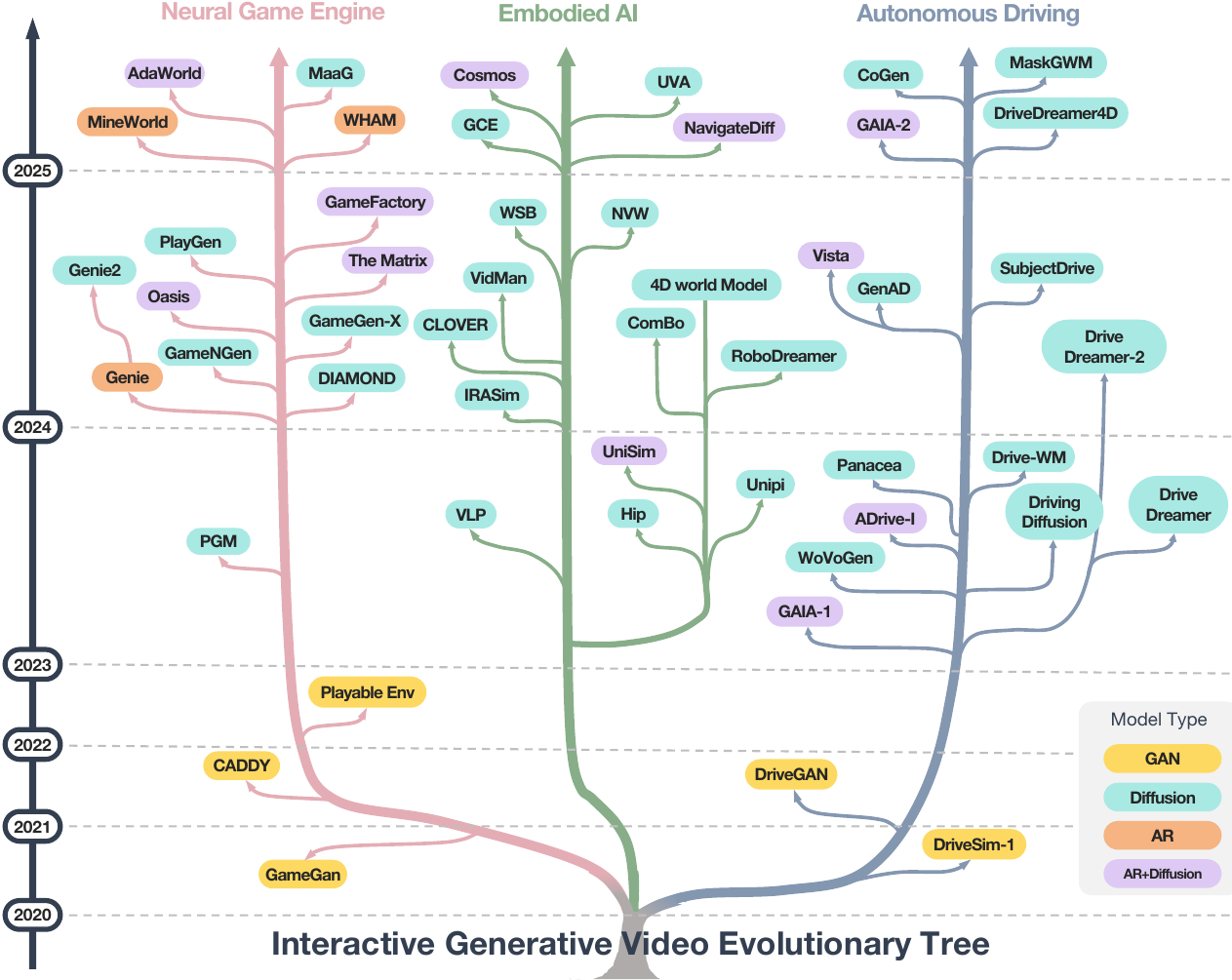}
  \caption{ Evolutionary tree of Interactive Generative Video (IGV) models from 2020 to 2025. This diagram categorizes the development of IGV research from three major application domains: Game Simulation, Embodied AI, and Autonomous Driving, each represented by a colored trunk. 
  }
\label{fig:motivation} 
\end{figure*}

Following our definition of IGV, we present in Fig.~\ref{fig:motivation} the evolution and development trajectory of three major IGV applications: Gaming, Embodied AI, and Autonomous Driving. In gaming applications~\cite{gamengine, gamegen-x,matrix,oasis,gamefactory,genie,genie2,wham,gamegan,caddy,pe,pgm,diamond,plan4mc,mineworld,maag,adaworld}, video games inherently combine visual output with player interaction, aligning perfectly with the core characteristics of IGV. IGV enables the creation of infinitely explorable interactive virtual worlds, where game content can be dynamically generated and personalized based on individual player preferences and skill levels. Moreover, IGV's generative capabilities significantly streamline game development by eliminating manual asset creation processes, substantially reducing development costs and improving efficiency. Notable examples include Oasis~\cite{oasis} and WHAM~\cite{wham}, both of which have released publicly playable versions that, although early-stage, demonstrate the preliminary exploration of IGV's potential in gaming.

In the field of embodied AI~\cite{vlp,hip,unipi,unisim,irasim,robodreamer,combo,worldsimbench,4dworldmodel,navigatediff,gce,cosmos,UVA}, IGV plays a crucial role in creating realistic and interactive simulations for robotic systems. It enables high-fidelity video sequence generation for task planning and visualization, allowing robots to better understand and interact with their environments. IGV also addresses the challenge of insufficient training data by providing diverse synthetic scenarios, thus improving policy learning and helping robots generalize across different tasks and environments.

For autonomous driving~\cite{drivegan,drivesim,gaia-1,wovogen,adriver,wen2024panacea,Drive-WM,drivingdiffusion,drivedreamer,drivedreamer-2,genad,vista,gaia-2,cogen,drivedreamer4d,maskgwm}, IGV offers advanced simulation capabilities that surpass traditional physics-based simulators in capturing real-world complexity. It generates high-fidelity video simulations conditioned on various control inputs, enabling comprehensive training across diverse driving scenarios. In addition, IGV enhances real-time decision making by predicting environmental changes and potential hazards, while also providing a safe platform to test autonomous systems in rare or dangerous situations.

While IGV has demonstrated promising applications across gaming, embodied AI, and autonomous driving, to better guide its future development in these domains, we propose a comprehensive framework that outlines the essential components of an ideal IGV system, as illustrated in Fig.~\ref{fig:igv_framework}. This framework helps identify key challenges and development directions through five fundamental modules:
The \textbf{Generation} module focuses on fundamental video generation capabilities; despite significant advances in generation quality, it still faces challenges in achieving real-time performance and frame-by-frame autoregressive generation. 
The \textbf{Control} module manages user interactions with the virtual world, with the primary challenge of enabling precise control while generalizing to open-domain scenarios. 
The \textbf{Memory} module ensures consistency in generated content across both static and dynamic aspects, struggling with maintaining long-term coherence. 
The \textbf{Dynamics} module simulates physical laws within the virtual world, with challenges in accurately simulating various physical phenomena and enabling precise tuning over specific physical parameters. 
The \textbf{Intelligence} module integrates causal reasoning capabilities, representing a higher level of intelligence that could potentially lead IGV to evolve into a self-evolving metaverse.

In this paper, we make three main contributions.
First, we provide a comprehensive survey of Interactive Generative Video applications and their current development status across various domains, including Gaming, Embodied AI, and Autonomous Driving.
Second, we propose a systematic framework that decomposes an ideal IGV system into five essential components, offering a structured approach to understand and develop IGV systems.
Finally, based on our proposed framework, we systematically analyze the technical challenges in realizing each component of an ideal IGV system, thereby providing clear directions for future research in this emerging field.

The structure of this survey is as follows. Sec.~\ref{sec:preliminary} presents the fundamentals technology of video generation, covering models like VAE, GAN, diffusion models, autoregressive models, and their hybrid forms. Sec.~\ref{sec:sys_overview} outlines the IGV system framework, including components such as Generation, Control, Memory, Dynamics, and Intelligence, and analyzes related challenges and future research directions.
Sec.~\ref{sec:app_video game} - \ref{sec:app_autodrive} are dedicated to IGV applications in gaming, embodied AI, and autonomous driving. Each application section describes the current situation, technical methods, challenges, and future development trends.

This work serves as an extended version of our previous research~\cite{igvposition}, with the following expansions: The framework we proposed for game engine extends beyond gaming technology requirements. It represents a trajectory towards higher intelligence in video generation models and will guide technological developments across multiple application domains. Specifically, we have supplemented our work with a comprehensive analysis of Interactive Generative Video (IGV) applications in broader fields, including Autonomous Driving and Embodied AI. Furthermore, we have identified corresponding challenges and proposed future research directions in these domains.

%% file: IGV_sections/2_preliminary.tex
\section{Preliminaries: Video Generation}\label{sec:preliminary}
\subsection{Overview} Video generation aims to synthesize realistic and temporally coherent video sequences. 
Advances in deep generative models have significantly influenced the development of video generation techniques. 
This section introduces these techniques, including a streamlined discussion on traditional models like Variational Autoencoders (VAEs)~\cite{kingma2019introduction} and Generative Adversarial Networks (GANs)~\cite{GANs}, and an expanded focus on models such as Diffusion models~\cite{ho2020denoising}, Flow matching methods~\cite{reflow,flowmatching} and Auto-Regressive (AR) approaches~\cite{xiong2024autoregressive}, and combined AR-Diffusion approaches~\cite{diffforcing,causaldiff}.

\subsection{Notations} 
To maintain consistency throughout this survey, we established a unified notation system. 
We denote a video sequence as $ \textbf{x} = \{x_0, x_1, \dots, x_t\} $,  where $x_i \in \mathbb{R}^{H \times W \times C}$ indicates the $i$-th frame image, where H and W represent the spatial dimensions (height and width), C represents the color channels.
The latent representation is denoted as $\mathbf{z} \in \mathbb{R}^d$, where d is the dimensionality of the latent space. 
We use $p(\mathbf{x})$ to represent the true data distribution of videos and $p(\mathbf{z})$ for the prior distribution of latent variables. 
Encoding and decoding processes are denoted by the encoder network $q_\phi(\mathbf{z}|\mathbf{x})$ and the decoder network $p_\theta(\mathbf{x}|\mathbf{z})$, respectively, with $\phi$ and $\theta$ representing network parameters. 
Random noise at timestep t is represented as $\epsilon_t$.

\begin{figure}[ht]
  \centering
  \includegraphics[width=0.9\linewidth]{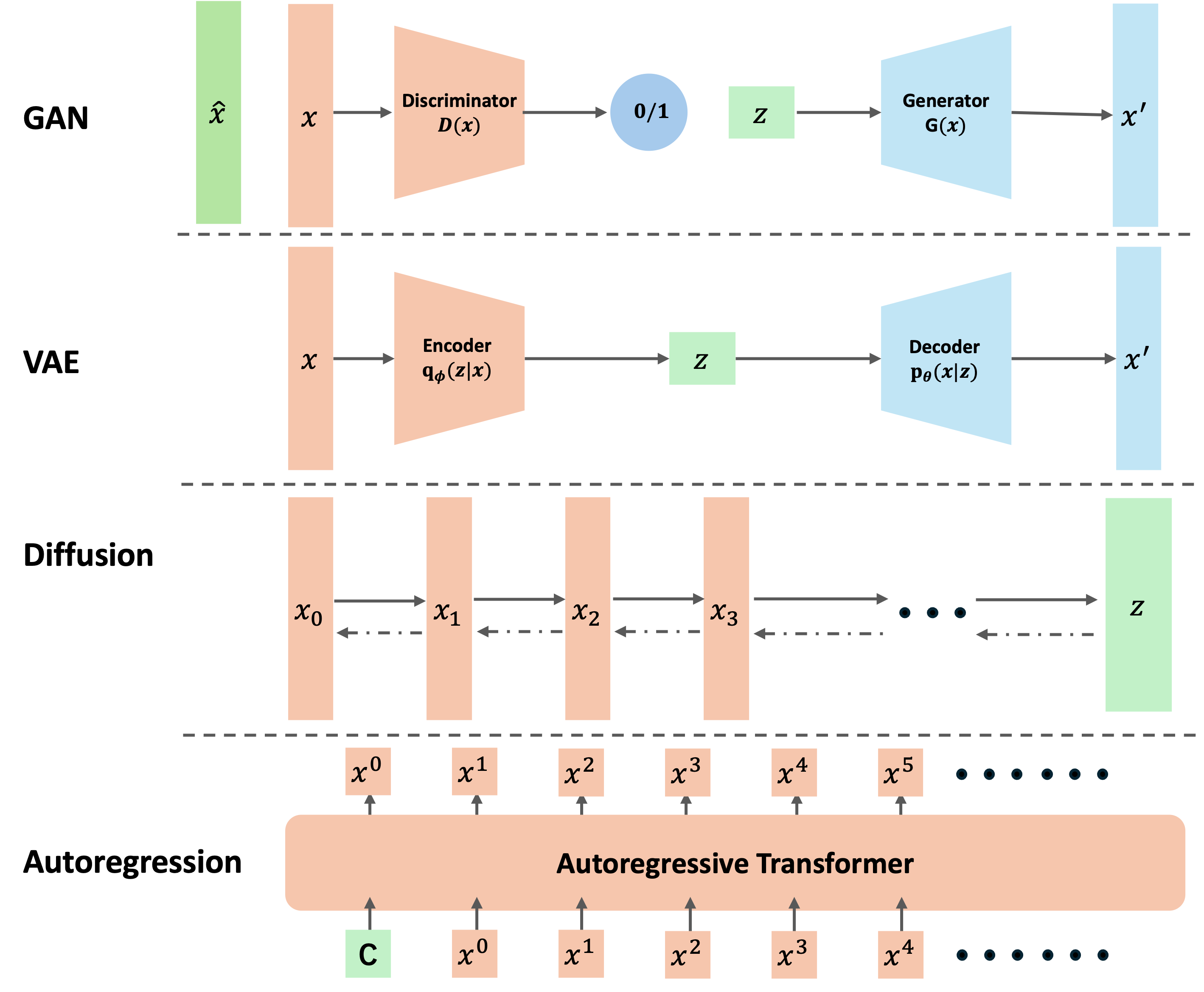}
  \caption{Overview figure for GAN, VAE, Diffusion, and Autogression methods. 
  }
\label{fig:pre} 
\end{figure}

\subsection{VAE and GAN} 
VAEs \cite{kingma2019introduction} and GANs~\cite{GANs} are classic methods for generation, which aim to learn a direct transformation between the latent space and video space. 
VAEs use a probabilistic framework to learn encoding and decoding processes, optimizing the evidence lower bound to balance reconstruction accuracy and latent space regularization via the Kullback-Leibler divergence. 
They optimize the evidence lower bound, expressed as $\mathcal{L}_{VAE} = \mathbb{E}_{q_\phi(\mathbf{z}|\mathbf{x})}[\log p_\theta(\mathbf{x}|\mathbf{z})] - D_{KL}(q_\phi(\mathbf{z}|\mathbf{x})||p(\mathbf{z}))$. 
The first term ensures accurate reconstruction, while the second term regularizes the latent space by minimizing the Kullback-Leibler divergence between the encoded distribution and a prior distribution. 
VAEs have been particularly successful in learning structured latent representations of videos, enabling controlled generation and manipulation.
GANs, through adversarial training, involve a generator and a discriminator to produce visually compelling video sequences by improving the generator's ability to fool the discriminator.
The objective function is formulated as $\min_G \max_D \mathbb{E}_{\mathbf{x}\sim p(\mathbf{x})}[\log D(\mathbf{x})] + \mathbb{E}_{\mathbf{z}\sim p(\mathbf{z})}[\log(1-D(G(\mathbf{z})))]$. 
The generator G learns to create increasingly realistic videos to fool the discriminator D, while D improves its ability to distinguish between real and generated videos.
This adversarial training mechanism has proven effective in producing sharp and visually compelling video sequences~\cite{tulyakov2018mocogan}.

\subsection{Diffusion Models}

Diffusion models~\cite{ho2020denoising} have become a leading framework for video generation due to their strong performance in fidelity and diversity. These models define a forward process that gradually adds Gaussian noise:
\[
q(\mathbf{x}_t|\mathbf{x}_{t-1}) = \mathcal{N}(\mathbf{x}_t; \sqrt{1-\beta_t}\mathbf{x}_{t-1}, \beta_t\mathbf{I}),
\]
and a reverse process that learns to denoise:
\[
p_\theta(\mathbf{x}_{t-1}|\mathbf{x}_t) = \mathcal{N}(\mathbf{x}_{t-1}; \mu_\theta(\mathbf{x}_t,t), \Sigma_\theta(\mathbf{x}_t,t)).
\]
Training typically minimizes the gap between the true noise and predicted noise:
\[
\mathcal{L}_{\text{diff}} = \mathbb{E}_{\mathbf{x}_0,\epsilon,t} \left[ \left\| \epsilon - \epsilon_\theta(\mathbf{x}_t,t) \right\|^2 \right].
\]

Early extensions of diffusion from image to video domains include VDM~\cite{ho2022video} and SVD~\cite{blattmann2023stable}, which introduce temporal consistency via 3D convolutions. More advanced models adopt spatio-temporal transformers~\cite{harvey2022flexible} or latent-space pipelines such as VideoFusion~\cite{luo2023videofusion}, VideoCrafter~\cite{chen2023videocrafter1,chen2024videocrafter2}, and LVDM~\cite{he2022latent}, which improve efficiency and resolution scalability.

Despite these advances, diffusion models remain computationally expensive due to their iterative sampling. To mitigate this, recent work explores \textbf{flow models}, which learn direct continuous mappings from prior to data without requiring multiple denoising steps. Flow Matching~\cite{flowmatching} learns a velocity field to transport samples:
\[
\mathcal{L}_{\text{FM}} = \mathbb{E}_{t, \mathbf{x}(t)} \left[ \left\| \mathbf{v}_\theta(\mathbf{x}(t), t) - \mathbf{v}^\star(\mathbf{x}(t), t) \right\|^2 \right],
\]
while Rectified Flow~\cite{reflow} simplifies the trajectory to a straight-line path with constant velocity. 
These approaches have shown promise in scaling to high-dimensional data such as video~\cite{opensora,kong2024hunyuanvideo,chen2025goku}.

Overall, diffusion and flow models continue to define the frontier of high-fidelity video generation, though efficiency and controllability remain active areas of improvement.

\subsection{Autoregressive Models}

Autoregressive models\cite{chen2020generative,yu2022scaling,team2024chameleon} generate video sequences by factorizing the joint distribution into conditionals:
\[
p(\mathbf{x}) = \prod_{t=1}^T p(\mathbf{x}_t|\mathbf{x}_{<t}),
\]
where each frame $\mathbf{x}_t$ is conditioned on all previous frames. This setup naturally captures temporal dependencies and aligns well with sequential data.

In practice, frames are often tokenized and processed using transformer-based language models. Notable examples include CogVideo~\cite{hong2022cogvideo}, which builds cascaded transformers over visual tokens, and VideoPoet~\cite{kondratyuk2023videopoet}, which adopts a unified latent space for multimodal generation. Emu3~\cite{wang2024emu3} further generalize this approach to support joint text-image-audio-video generation.

While autoregression provides strong temporal modeling, it suffers from two key limitations: (1) sequential decoding results in slow generation, and (2) early-frame errors can propagate and accumulate. These issues motivate hybrid frameworks that combine autoregression with other paradigms for improved efficiency and stability~\cite{wang2024loong,yan2021videogpt,wu2021godiva,ge2022long}.

\subsection{Hybrid AR-Diffusion Models}
\label{subsec:ar-diffusion}
To bridge the strengths of diffusion and AR models, recent studies explore hybrid frameworks—collectively referred to as \textit{AR+Diffusion}—that integrate sequential dependency modeling with high-fidelity generation. These methods can be broadly categorized along a spectrum, from AR-dominant to diffusion-dominant structures.

\textbf{(1) Autoregressive diffusion models (e.g., Diffusion Forcing~\cite{diffforcing}, CausalFusion~\cite{causaldiff}, AR-Diffusion\cite{sun2025ar}, DFoT~\cite{song2025history}, CausVid~\cite{yin2024slow}):} 
This category stays structurally closer to diffusion models but introduces temporal causality through autoregressive conditioning. Diffusion Forcing~\cite{diffforcing} applies non-uniform noise across frames to simulate AR-style information flow. CausalFusion~\cite{causaldiff} further imposes strict autoregressive conditioning during generation by factorizing video likelihoods, thus enforcing temporal coherence while retaining the multi-step denoising structure.

\textbf{(2) Decoupled AR and diffusion stages (e.g., Emu2~\cite{emu2}, SEED-X~\cite{ge2024seed}, GAIA-1\cite{gaia-1}):}  
These models are composed of an autoregressive model for multimodal prompt understanding, followed by a diffusion model for visual generation. The autoregressive multimodal LLM takes visual and textual prompts as inputs and predicts visual conditions for the diffusion model to generate images or videos. Training is typically conducted in multiple stages, by adapting pretrained multimodal LLMs for generating visual conditions, and aligning the input conditions of pretrained diffusion models.

\textbf{(3) Diffusion and AR with shared parameters (e.g., TransFusion~\cite{zhou2024transfusion}, Show-O~\cite{xie2024show}):}  
These approaches unify AR and diffusion objectives under a shared network architecture and parameterization. 
Typically, AR is used to model language tokens with next-token-prediction, and diffusion is adopted for visual generation. 
Despite the sharing of network architectures and parameters, the tasks of autoregression and diffusion are not unified.

\textbf{(4) AR backbone with lightweight diffusion sampler (e.g., MAR~\cite{mar}, NOVA~\cite{deng2024autoregressive}):}  
These methods leverage autoregressive models as the backbone for visual generation. In order to enable training and sampling with continuous tokens, they leverage a lightweight diffusion head to model the per-token distribution of each continuous token. Different from (2), this paradigm uses a lightweight diffusion head with only a few layers, and the AR module and diffusion head are trained end-to-end.

These hybrid approaches offer flexible trade-offs between generation speed, temporal consistency, and visual fidelity. Their design reflects the core demands of interactive generative video systems, where controllability, efficiency, and realism must be simultaneously optimized.

\input{fig_tab_subtex/igv_framework}

%% file: fig_tab_subtex/igv_framework.tex
\begin{figure*}[ht]
  \centering
  \includegraphics[width=0.9\linewidth]{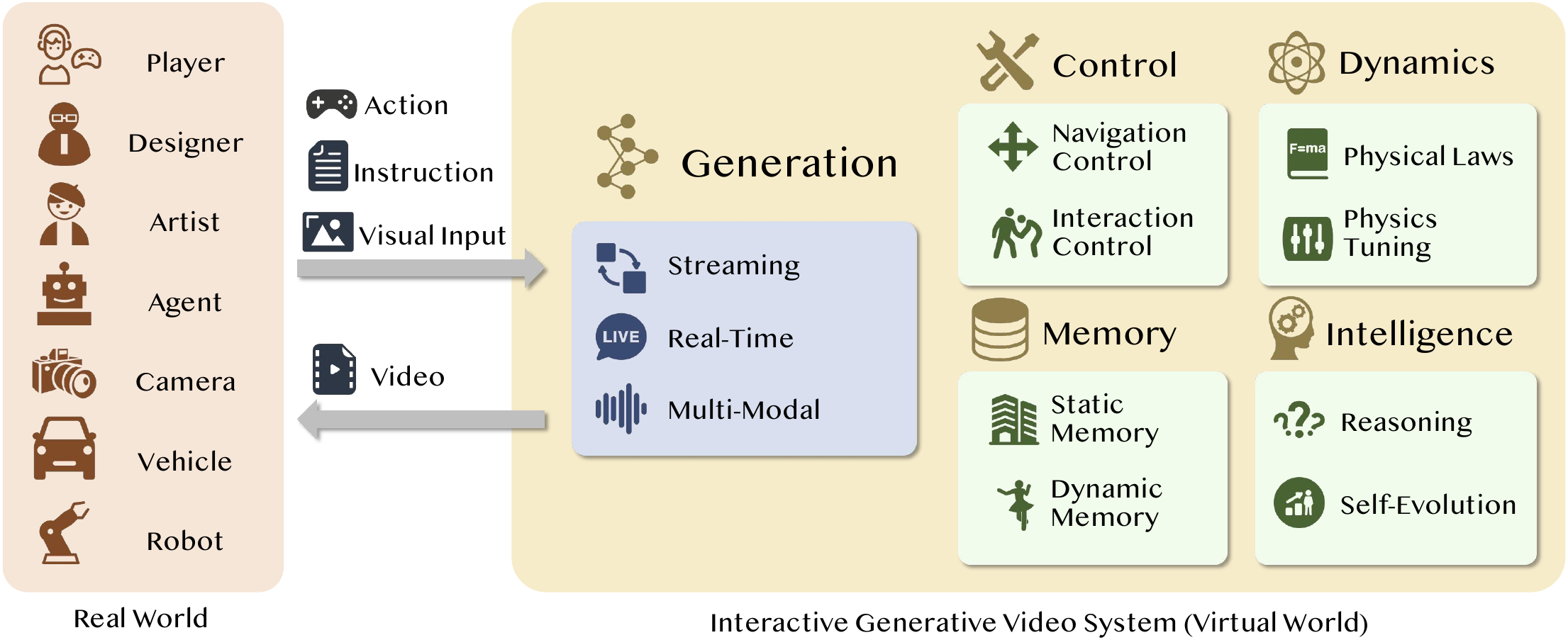}
  \caption{Proposed framework of Interactive Generative Video (IGV).
This figure illustrates the IGV system, which serves as a bridge between the real world and a virtual environment. In the real world, various roles such as players, designers, artists, and intelligent agents (e.g., robots, vehicles) interact with the IGV system through actions, instructions, and visual inputs. These diverse interactions naturally lead to applications in their respective domains: players engage with gaming applications, robots utilize embodied AI simulations, and vehicles operate in autonomous driving scenarios.
The IGV system generates video outputs through five interconnected modules: the Generation module serves as the foundation for video synthesis; the Control module enables precise manipulation of video content; the Memory module maintains content consistency; the Dynamics module ensures physical plausibility; and the Intelligence module represents higher-level capabilities like causal reasoning. These modules work in concert to create and manage immersive virtual experiences through video generation.
}
\label{fig:igv_framework} 
\end{figure*}

%% file: IGV_sections/3_sys_overview.tex
\section{IGV System}\label{sec:sys_overview}

In Figure \ref{fig:igv_framework}, we formulate the overall framework of Interactive Generative Video (IGV). The IGV system constitutes a virtual world which interfaces with various roles from the real world. These real-world roles include human actors, such as players, designers, and artists, who can interact with the IGV system to experience its virtual world or utilize it for efficient content creation. These roles also encompass various intelligent agents and their equipped sensors, such as robots, robotic arms, vehicles, and their mounted cameras, representing IGV's application potential in fields like autonomous driving and embodied intelligence. In the following subsections, we will introduce the key internal components of IGV and their technical challenges.

\subsection{Generation}
\subsubsection{Definition}
The Generation Module serves as the cornerstone of IGV systems, focusing on video synthesis capabilities. Beyond meeting fundamental requirements for visual fidelity and temporal consistency, this module incorporates three essential features to deliver superior interactive experiences:
\textbf{Streaming Generation} enables continuous frame-by-frame\footnote{While some methods like VAE-based temporal compression also perform frame-by-frame prediction, each "frame" in those cases represents a sequence of frame tokens corresponding to multiple actual frames.} video creation with precise temporal online control, allowing for theoretically infinite-length video generation. 
\exa{This capability supports endless procedural worlds, such as \textit{No Man's Sky}\footnote{\url{https://store.steampowered.com/app/275850/No_Mans_Sky/}}, where players can seamlessly explore for hundreds of hours. It also enables real-time environmental perception in autonomous vehicles for continuous processing of road conditions and traffic patterns. Furthermore, in rhythm games like \textit{Beat Saber}\footnote{\url{https://store.steampowered.com/app/620980/Beat_Saber/}}, this frame-by-frame generation ensures instant response to rapid user inputs where every frame matters}.
(2) \textbf{Real-time Processing} ensures instantaneous interaction with users.
\exa{This is essential in competitive games like \textit{Counter-Strike}\footnote{\url{https://store.steampowered.com/app/10/CounterStrike/}}, robotic surgery systems requiring precise instantaneous visual feedback, and \textit{League of Legends}\footnote{\url{https://www.leagueoflegends.com/}} where instant visual feedback is crucial}.
(3) \textbf{Multi-modal Generation} integrates video output with complementary modalities such as text and audio.
\exa{This includes dynamic music that responds to gameplay in \textit{Journey}\footnote{\url{https://store.steampowered.com/app/638230/Journey/}}, environmental awareness in embodied AI systems through multi-sensor fusion, and real-time dialogue subtitles in \textit{Mass Effect}}\footnote{\url{https://store.steampowered.com/app/17460/Mass_Effect_2007/}}.

\subsubsection{Challenges and Future Work}
\noindent (1) \textbf{Streaming Generation}

Diffusion-based methods have demonstrated exceptional capabilities in producing high-fidelity visual content. One straightforward implementation of streaming generation involves applying varying noise schedules across sequential frames~\cite{diffforcing,causaldiff,sun2025ar,song2025history,causvid}. Notable techniques such as diffusion forcing~\cite{diffforcing} and rolling diffusion~\cite{rolling} have witnessed various applications in video game content generation~\cite{thematrix,gamengine,oasis2024,gamefactory}. These methods enables sequential generation while maintaining visual fidelity, without necessitating architectural modifications to the underlying model.

Alternatively, next-token prediction presents a distinct pathway for autoregressive video synthesis~\cite{videopoet,emu3}. Although current implementations have not yet matched the visual standards set by diffusion methods, its potential for seamless integration with large language models, which suggests enhanced capabilities in causal reasoning, positions it as a compelling research direction~\cite{zhou2024transfusion,xie2024show}.

Recent innovations have explored hybrid architectures of diffusion models and next-token prediction mechanisms, aiming to preserve visual quality while improving temporal coherence~\cite{emu2, ge2024seed,gaia-1,zhou2024transfusion,xie2024show,mar, nova}, as introduced in Sec.~\ref{subsec:ar-diffusion}. Although these integrated approaches show initial promise, they are still in the nascent stages of development, and the potential to surpass established diffusion-based methodologies requires further investigation.

\noindent (2) \textbf{Real-time Generation}

Interactive video generation still faces significant computational overhead, making real-time performance a major challenge. From a computational efficiency point of view, performance optimization can be realized through several proven methodologies, including the compression of models through knowledge distillation~\cite{bk-sdm}, optimization of diffusion sampling procedures through ODE-based distillation techniques~\cite{videolcm}, implementation of efficient data compression through advanced VAEs or tokenization methods~\cite{dcae}, and architectural refinements focusing on causality to eliminate computational redundancy~\cite{causvid}. Although deployment-specific acceleration strategies play a crucial role in practical applications, they fall outside the primary focus of this analysis.

\noindent (3) \textbf{Multi-modal Generation}

While current interactive video generation primarily focuses on video modality output, the interaction medium between users and videos encompasses a broader spectrum of multi-modal information, such as subtitles and music, which remains largely unexplored.
A primary strategy involves developing comprehensive multi-modal architectures capable of processing and generating various data types, including textual content, visual information, acoustic signals, kinematic data, depth information, and neurological patterns. Contemporary research initiatives have begun exploring this integrated approach~\cite{zhou2024transfusion, xie2024show,videopoet,gato}, though substantial technical hurdles persist in achieving seamless integration.

An alternative methodology emphasizes the initial development of domain-specific large-scale models~\cite{pengi,depthanything,motiongpt,lvm} prior to their systematic integration into a cohesive framework. This approach necessitates careful consideration of inter-model relationships within the unified system architecture. For instance, language processing components generate detailed generation protocols, which subsequently guide video synthesis processes. These generated visual sequences then serve as input for audio synthesis modules, culminating in a synchronized multimodal output.

\subsection{Control}
\subsubsection{Definition}
The Control Module orchestrates user interaction with the virtual environment through two key dimensions:
(1) \textbf{navigation control} empowers users to traverse and discover the virtual space through camera positioning and character locomotion.
\exa{For example, in autonomous driving systems, operators use joystick controls for vehicle steering and speed adjustment, while monitoring multiple camera views for environmental awareness and trajectory planning.}
(2) \textbf{interaction control} facilitates user manipulation of virtual objects and elements. 
\exa{For instance, in robotic manipulation tasks, operators use haptic interfaces to control robotic arms, with force feedback for precise object grasping, and gesture recognition for intuitive control of end-effector orientation and position.}

\subsubsection{Challenges and Future Work}
The implementation architecture for control systems has been extensively researched. Predominant methodologies include: (1) Cross attention mechanisms~\cite{gamefactory,gamengine,matrix}, where navigational inputs are encoded into conditional feature representations functioning as keys and values, while video content features operate as queries. (2) External adaptation frameworks~\cite{gamegenx,motionctrl}, which facilitate direct integration of control and visual feature representations.

Although control mechanisms demonstrate reliability in constrained environments, their application must extend to broader, unrestricted domains. Several research initiatives~\cite{gamefactory,matrix,genie2} have utilized video generation foundations to address this challenge, yet the generalization of sophisticated interactions with limited control supervision remains a significant research challenge requiring further investigation.

The development of control systems, particularly in interactive contexts, extends beyond mechanical response patterns to encompass a comprehensive understanding of environmental interaction principles (as fundamental physical constraints). Contemporary research adopts a data-centric approach, focusing on the acquisition of comprehensive datasets across multiple domains: gaming datasets~\cite{gamefactory,gamegenx} for virtual world interactions, robotic manipulation trajectories~\cite{calvin,libero,robofactory} for embodied AI training, and diverse driving scenarios~\cite{nuscenes,waymo} for autonomous vehicle control.

In designing control interfaces, user intuition alignment serves as the fundamental principle across applications. In gaming, examples include gesture-based recognition systems and neural interfaces. In robotics, it involves creating adaptive control policies that seamlessly handle diverse manipulation tasks. For autonomous driving, it requires developing robust control strategies that balance safety constraints with natural driving behaviors. These developments collectively push towards more intuitive control mechanisms that better correspond to natural interaction patterns in their respective domains.

\subsection{Memory}
\subsubsection{Definition}
Existing video synthesis approaches primarily utilize attention mechanisms, exhibiting limitations in preserving scene compositions, object identities, and visual attributes during extended sequences or significant motion events. The Memory module tackles these limitations through two components: 
(1) \textbf{static memory} incorporates scene-level and object-level information retention, spanning game terrains, architectural structures, character assets, and object representations.
\exa{In autonomous driving systems, the module must consistently maintain the spatial layout of roads, buildings, and infrastructure; any inconsistency in the environmental mapping between frames could lead to critical navigation errors and safety risks.}
(2) \textbf{dynamic memory} manages temporal motion patterns and behavioral sequences, encompassing character animations, vehicle paths, particle systems, and dynamic environmental elements such as weather progressions. 
\exa{This is crucial in robotic manipulation tasks, where precise motion consistency is essential for the robot's end-effector movements and object interaction sequences, ensuring smooth and accurate transitions between different manipulation phases.}

\subsubsection{Challenges and Future Work}
Contemporary methodologies predominantly employ attention-based memory architectures, leveraging the inherent capacity of attention mechanisms to maintain temporal coherence through cross-attention operations between historical and generated frames~\cite{gamengine,oasis}. However, this approach demonstrates limitations in both memory retention fidelity and temporal scope constraints.

An alternative framework involves implementing specialized memory architectures, manifesting either as implicit high-dimensional feature representations~\cite{gamegan} or explicit three-dimensional structural encodings~\cite{pe,pgm,wonderjourney,viewcrafter}. These architectural components function as conditional parameters for the generation process, ensuring persistent maintenance of static environmental elements. While analogous approaches have shown promise in three-dimensional content generation~\cite{viewcrafter, see3d}, their adaptation for temporal video generation requires comprehensive exploration.

Dynamic memory systems, tasked with encoding temporal elements including animations, behavioral patterns, and motion trajectories, necessitate extensive dynamic video datasets, particularly those encompassing significant kinematic information such as human locomotion. The acquisition and annotation of high-fidelity dynamic content presents substantial challenges yet remains fundamental to advancing this research domain.

\subsection{Dynamics}
\subsubsection{Definition}
The Dynamics Module emphasizes two principal components:
(1) \textbf{Physical Laws} concentrates on interpreting and synthesizing videos that adhere to basic physical principles, particularly rigid body dynamics including gravitational forces, impact responses, and velocity changes.
\exa{In autonomous driving systems, precise physical modeling is crucial for predicting vehicle dynamics, braking distances, and collision avoidance, where accurate simulation of tire friction, momentum, and aerodynamics determines safe navigation.}
(2) \textbf{Physics Tuning} transcends \textbf{Physical Laws} by offering parametric control over physical properties rather than merely simulating real-world physics. This includes modifying gravitational constants, friction parameters, or directly adjusting temporal flow, motion rates, and object masses.
\exa{In robotic manipulation tasks, physics parameters can be dynamically adjusted to optimize performance, such as modifying grip forces for different materials, adjusting motion speeds based on object fragility, and fine-tuning joint dynamics for precise control.}

\subsubsection{Challenges and Future Work}
A fundamental data-driven methodology for physical law implementation involves learning probabilistic distributions from extensive video datasets to generate physically coherent outcomes~\cite{sora,cosmos}. However, this approach necessitates comprehensive high-fidelity video collections demonstrating diverse physical interactions, presenting a substantial challenge in contemporary research endeavors.

Physics-based memory control presents an alternative methodology, implementing direct physical simulations within memory architectures as conditional parameters for video synthesis, potentially yielding enhanced precision in physical behavior representation. One specific implementation utilizes video generation frameworks as rendering engines overlaid on physics simulation systems, ensuring precise physical compliance~\cite{motioncraft,physgen,physdreamer}. However, this approach remains constrained to physical phenomena that can be explicitly formulated within simulation frameworks.

The establishment of standardized evaluation metrics for assessing physical accuracy in generated content represents a critical research objective~\cite{worldsimbench}, facilitating the identification of limitations and guiding advancements in physical dynamics modeling methodologies.

The capability for physics parameter adjustment, though currently underexplored, represents a fundamental requirement for models to effectively comprehend and manipulate physical knowledge representations. The acquisition of synthetic video data with annotated physical parameters from simulation environments presents a potential solution pathway. This challenge is highlighted to stimulate future research initiatives in this domain.

\subsection{Intelligence}
\subsubsection{Definition}
The Intelligence Module encompasses two fundamental dimensions:
(1) \textbf{Reasoning}: This functionality facilitates extended temporal causality analysis from initial parameters, fostering richly interactive virtual environments.
\exa{For instance, the system can simulate how a nation's socioeconomic structure develops across generations based on its foundational resources and governing decisions, or model wildlife behavioral patterns in response to environmental shifts, such as fauna adapting their habitats when water sources become scarce. These mechanics are demonstrated in strategic simulation titles like \textit{Crusader Kings}\footnote{\url{https://store.steampowered.com/app/1158310/Crusader_Kings_III/}} and nature management simulations like \textit{Planet Zoo}\footnote{\url{https://store.steampowered.com/app/703080/_/}}.}
(2) \textbf{Self-Evolution}: This capability transcends basic video stream generation with dynamic virtual environments; it enables digital realms to progressively adapt, transform, and synthesize novel patterns, protocols, and interactions through emergent characteristics.
\exa{Within simulation environments, societies can naturally develop distinct cultural identities, biological systems can generate novel species variations, and urban landscapes can expand and transform organically. Such technological advancement points toward the possibility of creating immersive digital worlds similar to the movie \textit{The Matrix}, where numerous digital entities and participants exist within self-developing virtual environments.}

\subsubsection{Challenges and Future Work}
Analogous to the sophisticated causal reasoning capabilities exhibited by large language models, implementing reasoning functionalities in video generation systems necessitates the incorporation of causal architectural frameworks. This implementation requires conditioning new frame generation on historical video sequences through autoregressive mechanisms, complemented by extensive data-driven training to develop robust causal inference capabilities.

An alternative framework involves leveraging existing large language models or multimodal language architectures for causal reasoning in conjunction with video generation systems. This approach demonstrates particular promise in the development of unified comprehension and generation frameworks capable of performing both linguistic causal inference and visual content synthesis~\cite{zhou2024transfusion, xie2024show}. This integration represents a significant trajectory for future research initiatives.

Moreover, the successful integration and implementation of aforementioned capabilities—encompassing physics comprehension, physical simulation, and causal reasoning—could potentially catalyze the emergence of autonomous evolutionary capabilities. This convergence of advanced functionalities might facilitate the development of self-sustaining virtual environments, such as metaverse ecosystems populated by autonomous intelligent agents, or simulated reality constructs reminiscent of those portrayed in the movie \textit{The Matrix}.

%% file: IGV_sections/4_app1_game.tex
\section{Application\#1: IGV for Generative Video Game}\label{sec:app_video game}

\begin{figure}
    \centering
    \includegraphics[width=\linewidth]{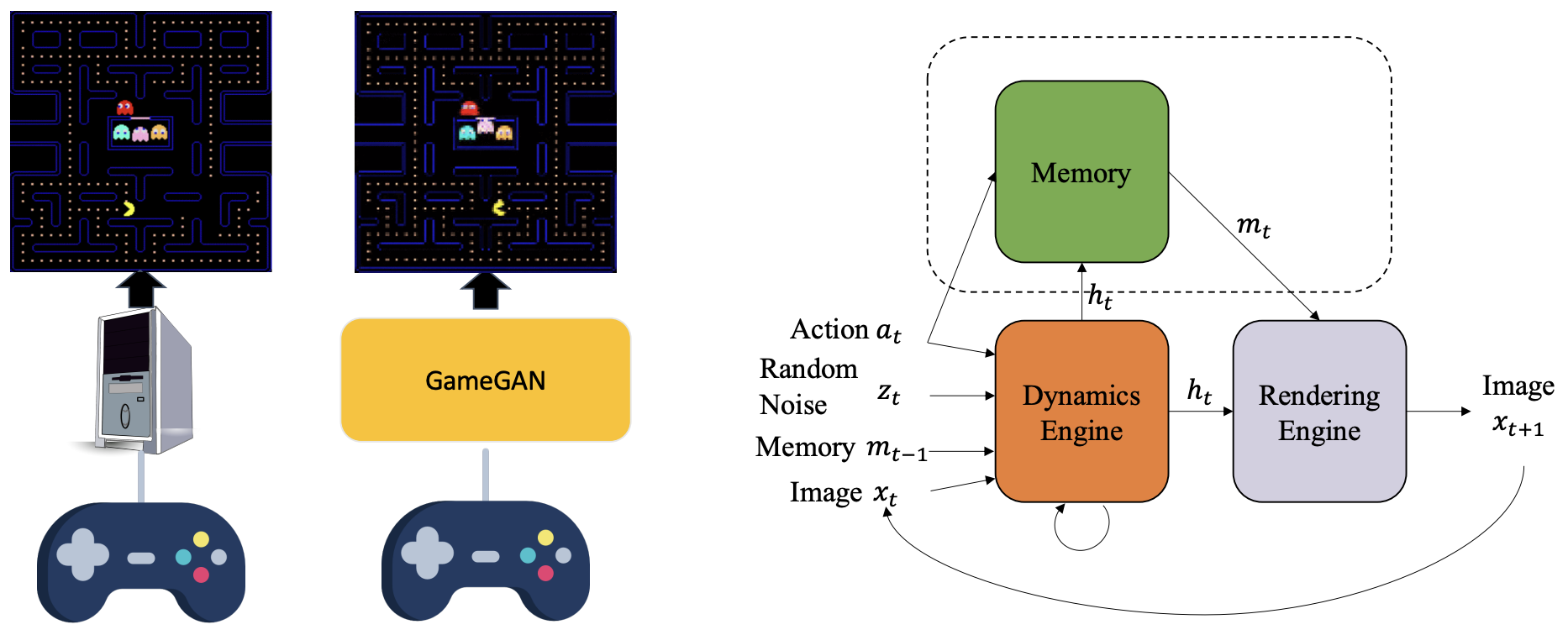}
    \caption{Overview figure of GameGAN~\cite{gamegan}. GameGAN is composed of three main modules. The dynamics engine (which refers to the internal mechanism that captures and updates the game's state transitions over time, simulating how the game world evolves in response to inputs~\cite{gamegan}) is implemented as an RNN and contains the world state updated at each time t. Optionally, it can write to and read from the external memory module. Finally, the rendering engine is used to decode the output image.}

    \label{fig:gamegan_vis}
\end{figure}

\begin{table*}[htbp]
  \centering
  \caption{\textbf{Overview of generative video models for generative video game.} Models are compared across modeling techniques, control signals, interaction type, real-time ability, creative generation, and open-domain control ability.}
  \resizebox{0.9\textwidth}{!}{%
  \begin{tabular}{lcccccc}
    \toprule
    Projects & Modeling Technique & Control Signal  & Interaction Type & Real-time & Creative  Generation & Open-domain Control \\
    \midrule
    World Model~\cite{worldmodel} & VAE & Embedding & Agent-based  & No & No & No \\
    GameGAN~\cite{gamegan} & GAN & Action & Direct  & Yes & No & No \\
    DriveGAN~\cite{drivegan} & GAN & Action & Direct  & No & No & No \\
    CADDY~\cite{caddy} & GAN & Action & Direct   & No & No & No \\
    Playable Env.~\cite{menapace2022playable} & GAN & Action & Direct   & No & No & No \\
    PGM~\cite{menapace2024promptable} & Diffusion & Action & Direct  & No & No & No \\
    DIAMOND~\cite{diamond} & Diffusion & Action & Agent-based  & No & No & No \\
    MarioVGG~\cite{MarioVGG} & Diffusion & Text & Direct  & No & No & No \\
    GameNGen~\cite{gamengine} & Diffusion & Action & Agent-based  & Yes & No & No \\
    Genie~\cite{genie} & Autoregression & Embedding & Agent-based  & Yes & Yes & Yes \\
    GenieRedux~\cite{kazemi2024learning} & Autoregression & Action & Agent-based & No & No & No \\
    Oasis~\cite{oasis2024} & Hybrid AR+Diffusion & Action & Direct  & Yes & No & No \\
    The Matrix~\cite{thematrix} & Hybrid AR+Diffusion & Text & Direct  & Yes & No & Yes \\
    PlayGen~\cite{yang2024playable} & Diffusion & Action & Agent-based  & No & No & No \\
    GameGen-X~\cite{gamegen-x} &Diffusion & Text/Action & Direct  & Yes & Yes & Yes \\
    GameFactory~\cite{gamefactory} & Hybrid AR+Diffusion & Action & Direct  & Yes & Yes & Yes \\
    Genie2~\cite{genie2} & Diffusion & Action & Direct  & Yes & Yes & Yes \\
    MineWorld~\cite{mineworld} & Autoregression &Action &Direct  &Yes &No &No\\
    AdaWorld~\cite{adaworld} & Hybrid AR+Diffusion &Embedding &Agent-based  &Yes &Yes &Yes\\
    MaaG~\cite{maag} &Diffusion &Action &Direct  &Yes &No &No\\
    WHAM~\cite{wham} &Autoregression &Action &Direct &No &No &No\\
    \bottomrule
  \end{tabular}%
  }
  \label{table_gvg}
\end{table*}
\subsection{Evolution of Generative Game Engine}

Generative Video Games are evolving into more sophisticated systems known as Generative Game Engines (GGE), which represent an emerging paradigm in interactive media where artificial intelligence systems dynamically create and control game environments through real-time player interactions~\cite{caddy,drivegan,diamond,gamegan,gamegen-x,gamengine,genie,kazemi2024learning,thematrix,worldmodel,yang2024playable,gamefactory,MarioVGG,oasis2024,genie2}. Defined as AI systems capable of generating persistent virtual worlds with responsive gameplay mechanics through continuous frame synthesis and state transition modeling, GGE fundamentally relies on IGV technologies to maintain temporal-spatial consistency while processing player inputs. This symbiotic relationship positions GGE as a specialized application domain of IGV, where the core challenges of maintaining visual-temporal coherence, implementing responsive game mechanics, and achieving real-time performance directly mirror IGV's technical requirements. In particular, IGV modules such as \textit{Generation}, \textit{Memory}, \textit{Control}, \textit{Dynamics}, and \textit{Intelligence} provide the underlying infrastructure for generative game development.

The field's trajectory reveals a paradigm shift from simulation to creation.
While early works like World Model~\cite{worldmodel} and GameGAN~\cite{gamegan} focused on replicating existing video games, recent advances exemplified by GameGen-X~\cite{gamegen-x} and Genie2~\cite{genie2} demonstrate the potential for open-domain video game content generation.
This evolution reflects a broader transition in AI: from mimicking existing patterns to creating novel, interactive experiences.
Additionally, the interaction framework in generative video games has evolved along dual paths: external interfaces (direct player control~\cite{gamegan,caddy,drivegan,gamegen-x,gamefactory,MarioVGG,oasis2024} versus agent-based interaction~\cite{diamond,gamengine,yang2024playable,kazemi2024learning}) with internal mechanics (from basic 2D operations~\cite{gamegan,diamond,genie,MarioVGG,worldmodel} to complex 3D world interactions\cite{oasis2024,gamengine,gamegen-x,gamefactory,thematrix,genie2}).
This dual-track development has enabled both immediate gameplay experiences and sophisticated AI-driven interactions, pushing the boundaries of what's possible in generated video game environments.
Technologically, the field has witnessed a crucial transition from VAE and GANs~\cite{gamegan,worldmodel,caddy,drivegan}to autoregressive or diffusion methods~\cite{diamond,gamengine,gamegen-x,gamefactory,oasis2024,thematrix,MarioVGG}, marked by significant improvements in generation quality and control.
The methods discussed in this section are summarized in Table~\ref{table_gvg}, which compares their modeling strategies and IGV-relevant capabilities. Here, \textit{Creative Generation} refers to a model's ability to produce novel game content beyond direct imitation, while \textit{Open-domain Control} indicates the generalization of in-domain control abilities to unseen data distribution.

The integration of IGV technology has been particularly transformative, providing essential frameworks for \textit{Autoregressive Generation}, \textit{Real-time Processing}, and \textit{Multi-modal Generation}.
These core capabilities enable continuous frame generation, tight user feedback loops, and multimodal interaction in real-time gameplay. The synergy points to a future where AI systems can create, control, and adapt video game environments on the fly, potentially revolutionizing how we think about video game design and player interaction. IGV not only advances technical capabilities but also opens new possibilities for creative expression in interactive digital environments.

\begin{figure}
    \centering
    \includegraphics[width=\linewidth]{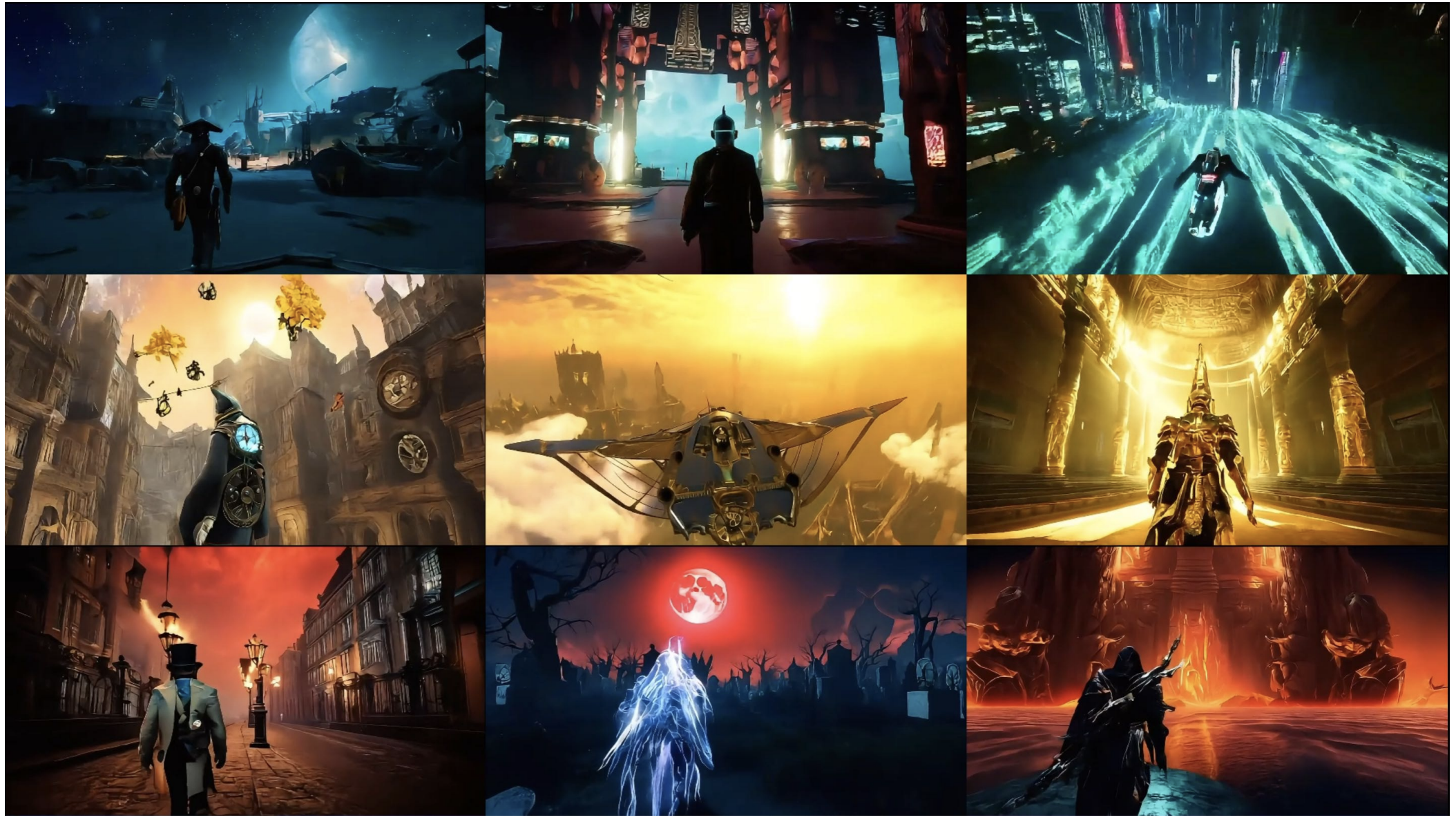}
    \caption{Open-domain generation showcases from~\cite{gamegen-x}.
    GameGen-X enables high-fidelity and diverse generation of open-domain video game scenes, supporting various styles, characters, and virtual environments with cinematic quality.}
    \label{fig:gamegenx_vis}
\end{figure}

\subsection{Technical IGV Approaches}
\subsubsection{From Video Game Simulation to Creation}
The evolution of GVG represents a paradigm shift in artificial intelligence, transitioning from passive video game simulation to active creative generation.
This transformation is driven by breakthroughs in neural architectures and training paradigms, enabling models not only to replicate existing game systems but also to raise novel gameplay mechanics, characters, and interactive manipulation.

\textbf{Phase 1: Foundational Work in Game Simulation (2018--2024):}
The field originated with seminal attempts to simulate game environments through neural networks~\cite{gamegan,diamond,gamengine,MarioVGG,oasis2024}.
The World Model~\cite{worldmodel} pioneered this direction by integrating RNN-based dynamics prediction and VAE-based rendering to simulate basic game physics and behaviors, demonstrating RL agents' learning potential in synthetic environments.
This inspired GameGAN~\cite{gamegan}, which established a framework mirroring IGV's modular design: a \textit{Memory Module} for state retention (analogous to IGV's Static and Dynamic Memory), a \textit{Dynamics Engine} for action-consequence modeling (mapping to the IGV Dynamics Module), and a \textit{Neural Renderer} that reflects IGV's \textit{Generation Module}. Its visual output, shown in Figure~\ref{fig:gamegan_vis}, demonstrates its ability to reproduce both classic 2D games and structured 3D-like environments.

\begin{figure}
    \centering
    \includegraphics[width=\linewidth]{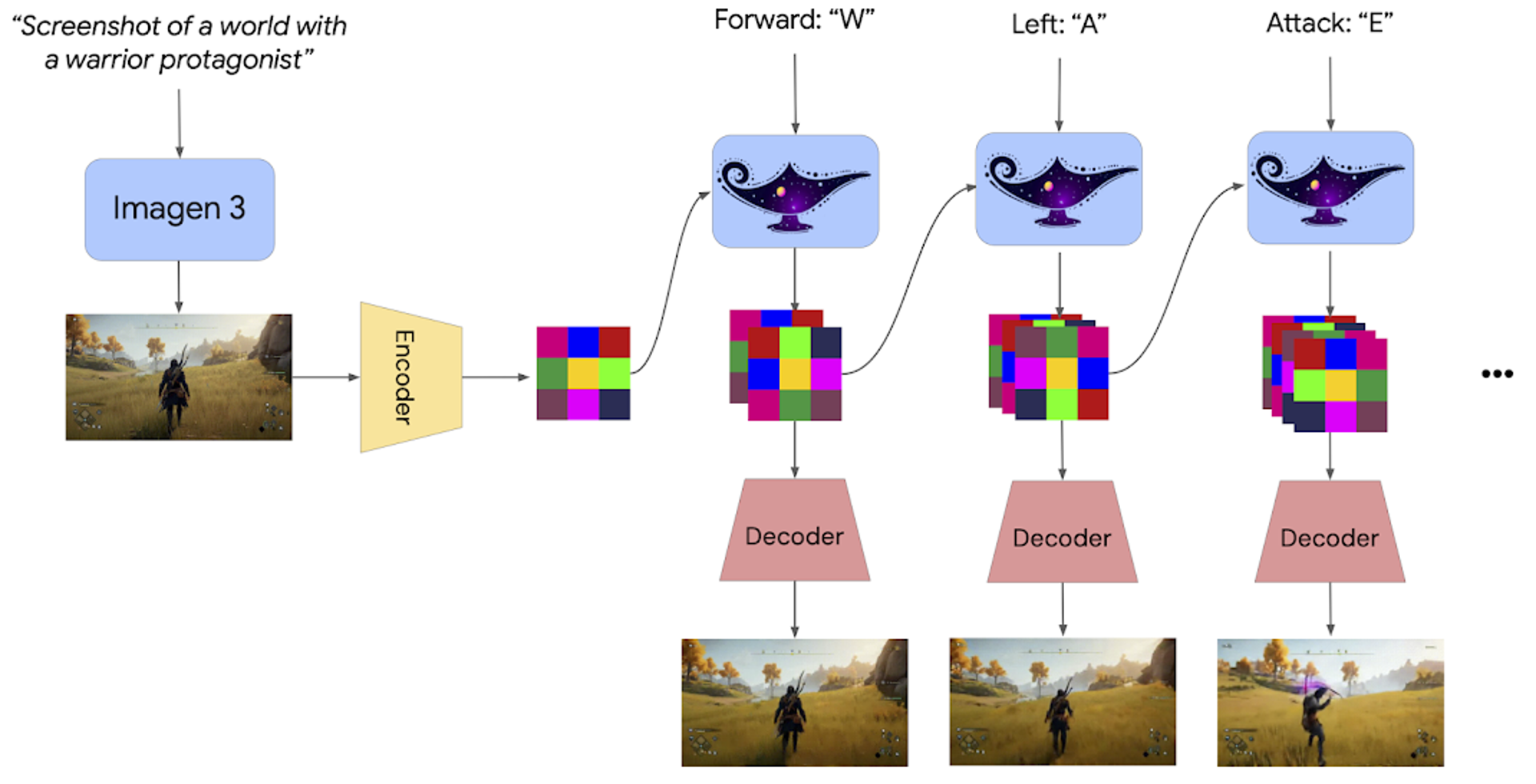}
    \caption{
    Overview figure of Genie2~\cite{genie2}.
    Genie2 is an autoregressive latent diffusion model. 
    At training time, latent frames are passed to a large transformer dynamics model, trained with a causal mask similar to that used by large language models.
    At inference time, Genie2 can be sampled in an autoregressive fashion, taking individual actions and past latent frames on a frame-by-frame basis. }
    \label{fig:genie2_vis}
\end{figure}

\textbf{Phase 2: Emergence of Open-Domain Generation (2024--Present):}
Recent advancements have transcended mere simulation, entering the era of \textit{creative generation} where models synthesize entirely original game content~\cite{gamegen-x,genie,genie2,gamefactory}.
This shift is enabled by three critical developments:  
\begin{itemize}
    \item \textbf{Scale-driven Emergence}: Genie~\cite{genie} demonstrated that unsupervised training on massive gameplay datasets allows models to learn latent action spaces and generate playable 2D games through trajectory prompting.
    \item \textbf{Multimodal Control:} GameGen-X~\cite{gamegen-x} established a full-stack framework for AAA-quality content generation using IGV’s \textit{Multi-modal Generation} capability, enabling text-to-character, sketch-to-environment, and physics-aware asset creation. The visual results in Figure~\ref{fig:gamegenx_vis} highlight its ability to generate richly varied, high-resolution game environments across diverse themes.
    \item \textbf{Frame-Level Agency:} Genie2~\cite{genie2} and Gamefactory~\cite{gamefactory} introduced temporal granularity and interaction grammar alignment, leveraging \textit{Autoregressive Generation} and \textit{Control Module} to enhance interactivity and playability. As illustrated in Figure~\ref{fig:genie2_vis}, Genie2 demonstrates real-time frame-level generation capabilities, tightly coupling user inputs with dynamic scene updates across a wide range of game settings.

\end{itemize}

\subsubsection{Interaction Dimensions in Generative Video Game}
The design and evolution of interaction systems in generative video games can be analyzed through two complementary analytical lenses: \textit{external interaction inference} (user-model engagement patterns) and \textit{internal interaction mechanisms} (model-world relationship processing). This dichotomy is embodied in the IGV \textit{Control Module}, which governs how player intentions are translated into game-world responses via \textit{navigation control} and \textit{interaction control}.

From the \textbf{external control signal} perspective, early models like GameGAN~\cite{gamegan} and DriveGAN~\cite{drivegan} rely on \textbf{direct action-based control}, where explicit user actions (e.g., keyboard input) are mapped to immediate game responses~\cite{oasis2024,thematrix}. 
In contrast, some systems, such as World Model~\cite{worldmodel}, Genie~\cite{genie}, and AdaWorld~\cite{adaworld} utilize \textbf{embedding-based control}, wherein control signals are inferred through latent representations, allowing higher-level abstraction of intent and semantic grounding.

The \textbf{interaction type} further differentiates how these control signals are interpreted. \textit{Direct interaction} systems tightly couple input and response, making them suitable for low-latency, reactive tasks but less flexible in semantic richness. In contrast, \textit{agent-based interaction} frameworks decouple control from execution, enabling models to reason about goals, memory, and temporal context. Systems like DIAMOND~\cite{diamond}, PlayGen~\cite{yang2024playable}, and GameNGen~\cite{gamengine} exemplify this agent-based design, integrating planning modules with perception and control layers.

\subsubsection{Trajectory of Model Architectures}
The evolution of generative modeling techniques in video games mirrors the trajectory of the IGV \textit{Generation Module}. Early systems such as World Model~\cite{worldmodel} and GameGAN~\cite{gamegan} employed VAE and GAN architectures, respectively, to map from a latent space or action input to output frames, offering basic frame synthesis but with limited control fidelity and semantic consistency.

Subsequent advances transitioned to \textbf{diffusion-based models} (e.g., DIAMOND~\cite{diamond}, GameNGen~\cite{gamengine}, GameGen-X~\cite{gamegen-x}), which emphasize \textit{conditional distribution modeling} to improve visual quality and enable fine-grained control over generation outputs. These models leverage noise-based iterative refinement, conditioned on text or action inputs, aligning closely with IGV’s goal of structured, high-fidelity generation. Similarly, GameFactory~\cite{gamefactory} extends game control priors to broader open-domain settings, enabling the reuse of handcrafted control logic originally designed for traditional games.

More recent frameworks incorporate \textbf{hybrid modeling techniques}, combining diffusion and autoregression. For example, Genie2~\cite{genie2} and AdaWorld~\cite{adaworld} fuse sequential autoregressive generation with diffusion refinement, supporting temporally coherent outputs across long horizons—an essential capability for interactive generative video (IGV) environments. 
Overall, this progression from implicit distribution modeling to explicitly conditional and hybrid formulations marks a maturation in generative model design, tightly coupling generation with interaction, memory, and control—core tenets of the IGV framework.

\subsection{Challenges and Future Directions}
\subsubsection{Consistency Optimization}
These consistency challenges are closely tied to IGV's \textit{Memory Module}, which addresses the preservation of visual and behavioral elements over time.
Future work can enhance \textbf{character consistency} by using identity-aware embeddings that encode pose, appearance, and interaction history.
\textbf{Scene consistency} can benefit from map-based structural rules and 3D Gaussian Splatting to improve spatial coherence.
\textbf{Historical consistency} may be reinforced via specialized encoders and dynamic scene graphs that track and update object relationships over time.

\subsubsection{Gameplay Enhancement}
Enhancing gameplay responsiveness requires the integration of IGV’s \textit{Control}, \textit{Dynamics}, and \textit{Real-time Generation} modules. Future work should improve rapid action feedback through high-frame-rate training, motion vector analysis, and real-time pose estimation. Developing unified frameworks that combine generation and understanding will better model complex scenarios. Expanding training datasets and incorporating dynamic reward systems can further enrich interaction diversity and player immersion.

\subsubsection{Real-time Performance Enhancement}
Real-time performance directly leverages IGV's \textit{Real-time Processing} capability.
Future work can explore lightweight diffusion models and autoregressive-nonautoregressive hybrids to balance quality and speed. Architectural optimizations, model distillation, and multi-stage refinement can reduce latency. Intelligent caching and adaptive quality control systems can further ensure consistent performance across hardware platforms.

\subsubsection{Toward Self-Evolving Games}
By incorporating IGV's \textit{Intelligence Module}, future generative video games may evolve into self-adaptive ecosystems. This includes learning emergent behavior patterns, adapting interaction rules, and generating new content without explicit supervision. With causal reasoning, physical understanding, and memory evolution, generative games may eventually form persistent, evolving virtual worlds akin to \textit{The Matrix} or open-ended metaverses.

%% file: IGV_sections/5_app2_embodied.tex











\section{Application\#2: IGV for Embodied AI}\label{sec:app_embodied}

\begin{figure}
    \centering
    \includegraphics[width=1.0\linewidth]{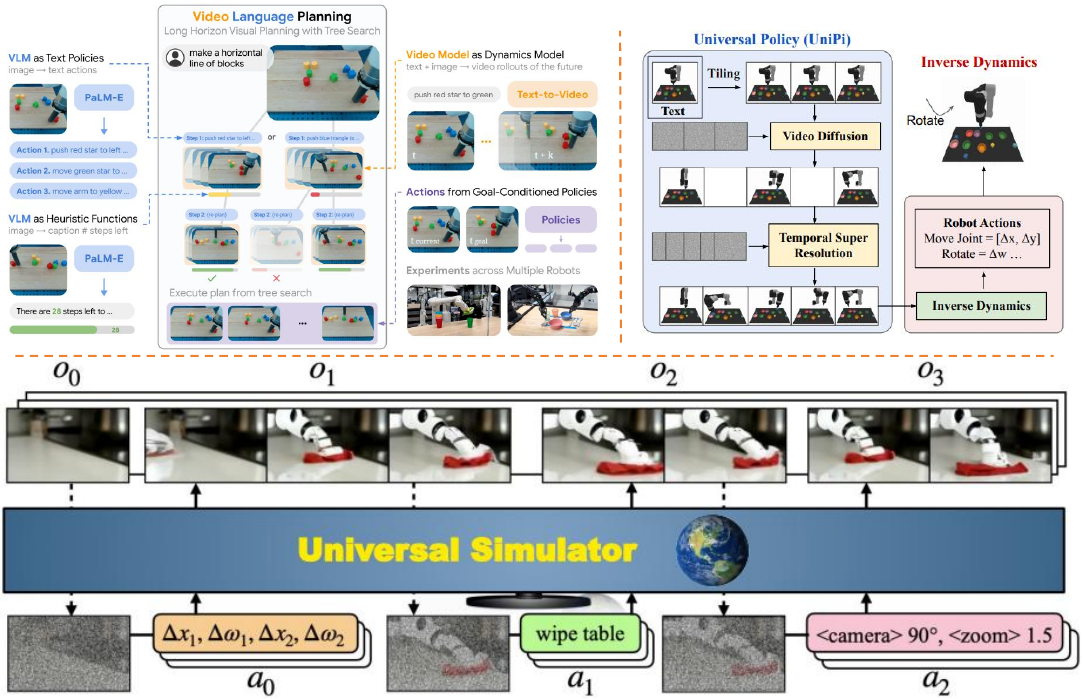}
    \caption{Overview of three applications of IGV in embodied intelligence. \textbf{Top-left:} VLP~\cite{vlp} demonstrates the use of IGV for embodied planning. \textbf{Top-right:} UniPi~\cite{unipi} illustrates how IGV can serve as a generalizable robot policy. \textbf{Bottom:} UniSim~\cite{unisim} showcases the potential of IGV as a world simulator.}

    \label{fig:eai}
\end{figure}

\begin{table*}[htbp]
  \centering
  \caption{\textbf{Overview of generative video models for Embodied AI.} Models are compared across tasks, modeling paradigms, control signals, and output modalities, highlighting their applicability to IGV-based embodied learning.}
  \label{tab:app_eai}
  \resizebox{\textwidth}{!}{%
  \begin{tabular}{lcccccc}
    \toprule
    Projects & Application & Specific Task & Modeling Technique & Control Signal  & Output & Dataset   \\
    \midrule
    VLP~\cite{vlp} & Task Planning & Manipulation & Diffusion & Image,Text & Img & Multiple data sources~\cite{bridge,rt-2,ego4d,epic-kitchens,Laion-400m} \\
    COMBO~\cite{combo} & Task Planning & Manipulation & Diffusion & Img,Text & Img & COMBO~\cite{combo}  \\
    Genex~\cite{genex} & Task Planning & Navigation & Diffusion & Img,Text & Img & Genex~\cite{genex}  \\
    IRASim~\cite{irasim} & Policy Learning & Manipulation & Diffusion  & Img,Trajectory & Img & Multiple data sources~\cite{bridge,rt-1,Language-Table}  \\
    UniSim~\cite{unisim} & Policy Learning & Manipulation,Navigation & Hybrid AR+Diffusion & Img,Text,Motor Control,Act & Img & Multiple data sources~\cite{habitat,hm3d,Language-Table,bridge,rt-1,rt-2,ego4d,epic-kitchens,something2something,matterport3d,Laion-400m}  \\
    NWM~\cite{nwm} & Policy Learning & Navigation & Diffusion & Img,Traj & Img & Multiple data sources~\cite{SCAND,tartandrive,RECON,HuRoN,ego4d}  \\
    Cosmos~\cite{cosmos} & Policy Learning & Manipulation, Navigation & Hybrid AR+Diffusion & Img,Text,Traj, Control,Act & Img & Multiple data sources~\cite{cosmos}  \\
    Unipi~\cite{unipi} & Video Action Learning & Manipulation & Diffusion & Img,Text & Img,Act & PDSketch~\cite{pdsketch}  \\
    Susie~\cite{susie} & Video Action Learning & Manipulation & Diffusion & Img,Text & Img,Act & Calvin~\cite{calvin}  \\
    HiP~\cite{hip} & Video Action Learning & Manipulation & Diffusion & Img,Text & Img,Act & Multiple data sources~\cite{pdsketch,cliport,kitchenshift} \\
    RoboDreamer~\cite{robodreamer} & Video Action Learning & Manipulation & Diffusion & Img,Text & Img,Act & RT-1  \\
    DreamerV3~\cite{dreamerv3} & Video Action Learning & Manipulation,Navigation & CNN & Img,Act & Img,Act & MineRL~\cite{minerl}  \\
    MineDreamer~\cite{minedreamer} & Video Action Learning & Manipulation,Navigation & Hybrid AR+Diffusion & Img,Text & Img,Act & MineRL~\cite{minerl}  \\
    WSB~\cite{worldsimbench} & Video Action Learning & Manipulation,Navigation & Diffusion & Img,Text & Img,Act & Multiple data sources~\cite{minerl,dosovitskiy2017carla,calvin} \\
    4D WorldModel~\cite{4dworldmodel} & Video Action Learning & Manipulation & Diffusion & Img,Text & Img,Depth,Normal,Act & RLBench~\cite{rlbench},RT-1~\cite{rt-1}  \\
    NavigateDiff~\cite{navigatediff} & Video Action Learning & Navigation & Hybrid AR+Diffusion & Img,Text & Img,Act & Gibson~\cite{gibson}  \\
    GCE~\cite{gce} & Video Action Learning & Manipulation,Navigation & Diffusion & Img,Text & Img,Act & Multiple data sources~\cite{libero,metaworld,calvin,ithor} \\
    UVA~\cite{UVA} & Video Action Learning & Manipulation & Diffusion & Img,Text & Img,Act & UMI~\cite{umi}  \\
    \bottomrule
  \end{tabular}%
  }
\end{table*}
\subsection{Learning-Based Robotic Tasks}

Embodied AI aims to enable agents (such as robots) to learn and perform tasks in the physical world, from manipulating objects to navigating through complex environments. These learning-based robotic tasks pose formidable challenges. Robotic manipulation requires fine-grained control of contact dynamics and understanding of varied objects, while navigation demands interpreting 3D scenes and long-horizon planning. Both domains involve high-dimensional sensory inputs and continuous action spaces under partial observability, making it difficult to hand-code solutions. Moreover, environments can differ vastly between tasks – each home, factory, or lab setting has distinct state and action spaces, hampering the transfer of knowledge and policies across scenarios. Designing reward functions or goal specifications that generalize is equally problematic, since success criteria vary with each task. 

Recent advances suggest that generative models can help tackle these issues by synthesizing realistic data and simulating futures for decision-making. In robotic manipulation, for instance, generative models are used to create synthetic training examples, plan feasible trajectories, and adapt to unstructured settings. By capturing underlying patterns in vast datasets, such models enhance a robot’s ability to generalize, make decisions, and execute tasks in dynamic environments. Interactive Generative Video (IGV) is proposed as a unifying framework to leverage these strengths in service of embodied AI. IGV extends state-of-the-art video generation with interactivity, effectively functioning as a learnable simulator for the real world~\cite{unisim}. 

By providing unlimited, diverse, and realistic experiences through this video-based virtual world, IGV can support learning-based robotics in ways traditional simulators or static datasets cannot. An embodied agent can train and evaluate its skills within IGV’s generated scenarios, benefiting from high-fidelity sensory inputs and responsive feedback that adhere to real-world constraints. In the following, we discuss three key capabilities for embodied AI which shown in the Tab.~\ref{tab:app_eai} and Fig.~\ref{fig:eai} and how IGV enhances each of them: (1) long-horizon task planning, (2) policy learning from interaction, and (3) generalization across tasks and environments.

\subsection{Generative Video for Task Planning}
Planning in embodied AI involves translating high-level goals (e.g., “tidy the room”) into actionable low-level behaviors. This process requires understanding semantic intent, reasoning over environment dynamics, and generating physically feasible action sequences — challenges that traditional planners struggle with in open, dynamic environments. 

Traditional robot planning methods often rely on hand-crafted rules and language models~\cite{vima,code-as-policy,robofactory,voxposer,rekep}, but these methods are limited when dealing with complex and dynamic environments, especially with continuous visual input. IGV offers a powerful solution by leveraging its Generation and Dynamics modules to simulate future outcomes as high-fidelity video sequences, effectively serving as a forward world model. The Intelligence module supports high-level reasoning, while Memory ensures temporal and spatial consistency across generated scenes.

Recent works such as VLP~\cite{vlp} introduces an innovative approach for visual planning of complex, long-horizon tasks using large generative models pre-trained on vast Internet-scale data. VLP combines IGV to create detailed multimodal video plans that guide robots through tasks, from multi-object rearrangement to bi-arm dexterous manipulation, by generating step-by-step videos and corresponding textual instructions. 
COMBO~\cite{combo} proposed a world model approach for multi-agent cooperation using IGV, where the world model is decomposed into sub-models and supported by generative models to create Interactive Generative Videos for complex task coordination and planning. 
Genex~\cite{genex} introduces Generative World Explorer which leverage panoramic video generation to generate imagined observations and update their beliefs about large-scale 3D environments like urban scenes without physical exploration, helping agents make more informed plans based on the updated beliefs.

These approaches demonstrate how IGV enables robots to ``imagine'' and validate plans visually, significantly improving planning robustness and flexibility in real-world and simulated environments.

\subsection{Policy Learning via Generative Simulation}

To improve robot performance in complex tasks, traditional data collection methods often fail to provide sufficient diversity and quantity of data. While traditional policy learning relies on simulation environments, these simulations often have limitations, such as approximating reality and omitting subtle physical dynamics, which can result in a “reality gap” where policies that succeed in simulation fail in the real world. 
IGV offers a safer and more scalable alternative by leveraging synthesized data, particularly generated through IGV itself, allowing robots to undergo large-scale training in virtual environments. This method provides robots with rich and varied data while reducing real-world interactions, improving policy training efficiency and safety. By simulating realistic visual experiences with physical consistency through its Generation, Control, and Dynamics modules, IGV enables robots to train policies in a closed-loop setup, mimicking real sensor-action feedback, thereby bridging the gap between simulation and reality.

IRASim~\cite{irasim} introduces an interactive world model that utilizes generative models to create realistic video simulations of robot arms performing actions from a specified initial frame that helps robots learn and plan more efficiently by simulating interactions with the real world. 
Additionally, UniSim~\cite{unisim} proposed a generative model designed to simulate realistic interactions within virtual environments, responding to both high-level instructions and precise control inputs. The large amount of robot data generated by UniSim can effectively assist robot policy learning and reduce the cost of data collection.

Altogether, IGV transforms policy learning into a data-rich, visual imagination process, enabling robust skill acquisition without extensive real-world interaction.

\subsection{Generalization across Tasks and Environments}

A core challenge in embodied AI is generalization—enabling agents to adapt to unseen tasks, objects, and environments. IGV supports this through its Generation and Memory modules, which allow compositional scene construction and the synthesis of out-of-distribution scenarios, helping agents avoid overfitting to narrow training distributions.

Unipi~\cite{unipi} proposed a method for generating task-specific interactive videos through text guidance, thereby training universal control policies. 
HiP~\cite{hip} explored how to use IGV to generate tasks at different levels of hierarchy and achieve task decomposition and composition by combining these generated videos. 
Moreover, RoboDreamer~\cite{robodreamer} addresses the generalization issue common in existing models by factorizing video generation based on the compositional nature of language, breaking down instructions into actionable primitives, and facilitating compositional generalization in robot tasks.
\cite{4dworldmodel} use 4d Gaussian reconstruction~\cite{gs} to reconstruct dynamic 4d scenes, provide accurate depth and geometric information, assist interactive video generation, and improve the accuracy of robot control.

Together, these approaches showcase IGV’s potential to foster robust generalization by enabling imagination, composition, and abstraction across tasks and domains.

\subsection{Challenges and Future Work}

Interactive Generative Videos (IGV) offer transformative potential for embodied AI applications such as task planning, policy learning, and generalization. However, practical deployment of these systems still faces several critical limitations. Below, we identify and discuss three key challenges aligned with the previous sections and outline promising directions for future research.

\subsubsection{Challenge \#1: Planning with IGV under Real-Time and Complex Conditions}
\textbf{Related to:} \textit{Generative Video for Task Planning}

While IGV provides a powerful mechanism for visualizing future action outcomes and aiding task planning, real-time generation of high-quality, interactive video remains a major bottleneck—particularly in scenarios requiring immediate responses or rapid re-planning. Furthermore, as planning scales to multi-agent settings or complex manipulation pipelines, maintaining temporal coherence and computational efficiency becomes increasingly difficult.

\textbf{Future direction:} Research should focus on improving the efficiency of generative architectures, such as through lightweight diffusion or autoregressive video models, and enhancing temporal planning fidelity via hierarchical or hybrid symbolic-generative planners. Parallel work on scalable simulation management and real-time IGV deployment across heterogeneous robot fleets will also be essential.

\subsubsection{Challenge \#2: Bridging Sim-to-Real Gaps in Policy Learning with IGV}
\textbf{Related to:} \textit{Policy Learning via Generative Simulation}

Although IGV allows agents to train on rich and diverse synthetic data, ensuring that this data captures the full spectrum of real-world variability is challenging. Traditional simulators already suffer from the "reality gap," and IGV, while more expressive, may still generate unrealistic dynamics or visual artifacts if not properly constrained. Furthermore, policies trained in IGV environments may fail to generalize if sim-to-real transfer is not robustly addressed.

\textbf{Future direction:} Enhancing the realism and domain diversity of IGV-generated data is critical. This includes integrating advanced domain randomization, real-world fine-tuning, and curriculum learning. Additionally, incorporating domain adaptation or sim-to-real transfer techniques—such as adversarial training, style transfer, or real-to-sim bridging models—will help make IGV-based policy learning more applicable to physical systems.

\subsubsection{Challenge \#3: Generalization and Multimodal Integration in Universal Policies with IGV}
\textbf{Related to:} \textit{Generalization across Tasks and Environments}

Designing universal policies that generalize across unseen tasks, objects, and environments remains an open challenge. While IGV enables compositional video generation and abstraction through language or visual conditioning, it remains difficult to scale such models to arbitrary combinations of modalities, instructions, and interaction types. Additionally, many IGV models struggle to represent fine-grained task variations or to perform hierarchical reasoning across modalities.

\textbf{Future direction:} Future research should focus on enhancing the generalization ability of IGV systems using techniques such as meta-learning, few-shot adaptation, and instruction tuning. Moreover, improving multimodal fusion—especially for visual-textual-auditory inputs—can empower IGV to act as a unified policy interface. Building IGV models that natively reason over cross-modal input and support modular, reusable video segments will be key to developing versatile, goal-driven robotic agents.


%% file: IGV_sections/6_app3_autodrive.tex
\section{Application\#3: IGV for Autonomous Driving}\label{sec:app_autodrive}

\begin{figure}
    \centering
    \includegraphics[width=1.0\linewidth]{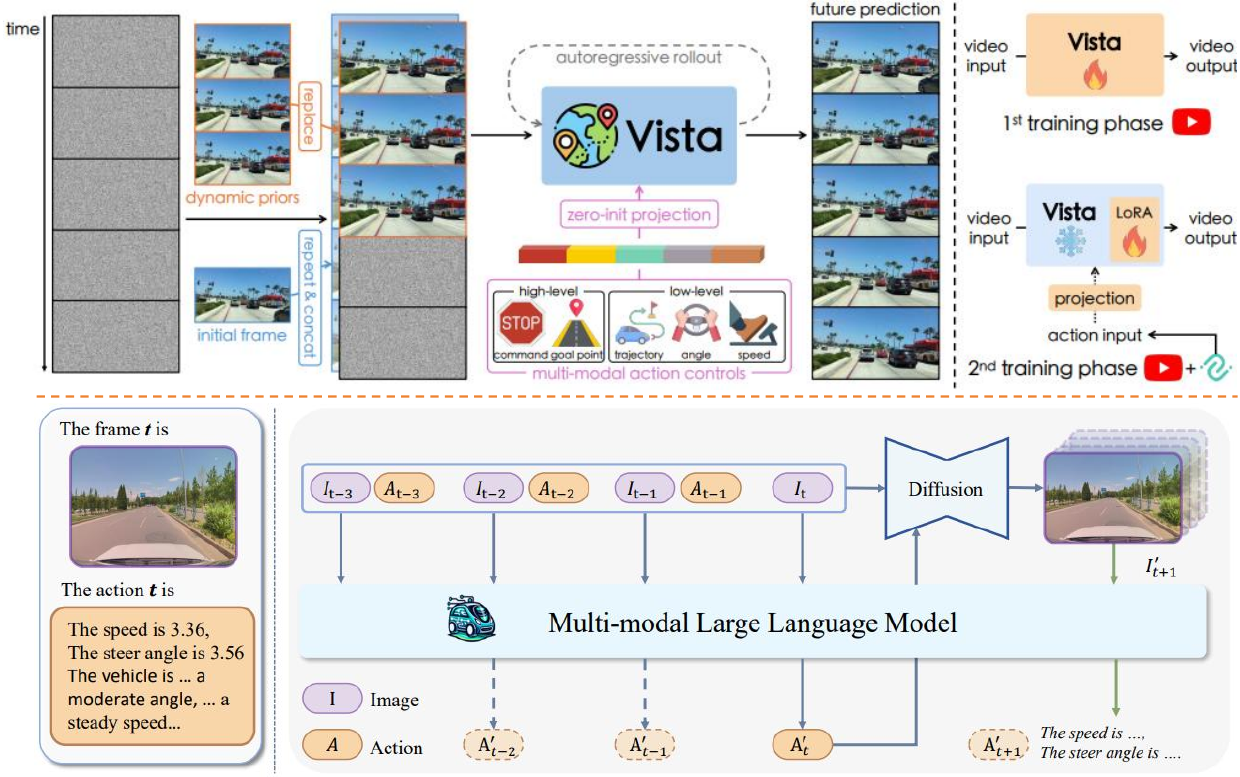}
    \caption{Overview of two applications of IGV in autonomous driving. \textbf{Top:} Vista~\cite{vista} demonstrates the potential of using IGV as a simulator for autonomous driving. \textbf{Bottom:} ADriver-I~\cite{adriver} illustrates the use of IGV as an interpretable vehicle controller.}

    \label{fig:ad}
\end{figure}

\begin{table*}[htbp]
  \centering
  \caption{\textbf{Overview of generative video models in driving simulation within Embodied AI.} Models are compared in terms of their generative techniques, control signal integration, output formats, and datasets, highlighting their relevance to Interactive Video Generation (IGV) tasks.}
  \label{tab:app_ad}
  \resizebox{\textwidth}{!}{%
  \begin{tabular}{lccccc}
    \toprule
    Projects & Application & Modeling Technique & Control Signal  & Output & Dataset   \\
    \midrule
    DriveSim~\cite{drivesim} & Driving Simulation & GAN & Img & Img & DriveSim~\cite{drivesim}  \\
    DriveGAN~\cite{drivegan} & Driving Simulation  & GAN & Img, Act & Img & Carla~\cite{dosovitskiy2017carla}, Real World Data~\cite{drivegan}  \\
    GAIA-1~\cite{gaia-1} & Driving Simulation & Hybrid AR+Diffusion & Img,Text,Act & Img & Wayve~\cite{gaia-1}   \\
    WoVoGen~\cite{wovogen} & Driving Simulation & Diffusion & Img, Text, HD, Occ, Obj & Img, HD, Occ & nuScenes~\cite{nuscenes}  \\
    Panacea~\cite{wen2024panacea} & Driving Simulation  & Diffusion & Img, Text, BEV Seq  & Img & nuScenes~\cite{nuscenes}  \\
    DrivingDiffusion~\cite{drivingdiffusion} & Driving Simulation  & Diffusion & Img, Flow, Text, 3D Lay & Img & nuScenes~\cite{nuscenes}  \\
    DriveDreamer-2~\cite{drivedreamer-2} & Driving Simulation  & Diffusion  & Img, Text, HD & Img & nuScenes~\cite{nuscenes}  \\
    GenAD~\cite{genad} & Driving Simulation  & Diffusion & Img, Text, Act & Img  & OpenDV-2K,nuScenes  \\
    Vista~\cite{vista} & Driving Simulation  & Hybrid AR+Diffusion & Img, Text, Act & Img  & OpenDV-2K~\cite{opendv2k},nuScenes~\cite{nuscenes}  \\
    GAIA-2~\cite{gaia-2} & Driving Simulation  & Hybrid AR+Diffusion & Img, Act & Img  & GAIA-2~\cite{gaia-2}  \\
    CoGen~\cite{cogen} & Driving Simulation  & Diffusion & Img, Layout & Img  & nuScenes~\cite{nuscenes}  \\
    DriveDreamer4D~\cite{drivedreamer4d} & Driving Simulation  & Diffusion & Img, Traj, Layout & Img  & Waymo~\cite{waymo}  \\
    MaskGWM~\cite{maskgwm} & Driving Simulation  & Diffusion & Img, Act & Img  & OpenDV-2K, nuScenes, Waymo~\cite{waymo}  \\
    Drive-WM~\cite{Drive-WM} & Video Action Learning  & Diffusion & Img, Text, BEV Seq  & Img, Traj & nuScenes~\cite{nuscenes}  \\
    DriveDreamer~\cite{drivedreamer} & Video Action Learning  & Diffusion  & Img, Act, Box, Text & Img, Act & nuScenes~\cite{nuscenes}  \\
    ADriver-I~\cite{adriver} & Video Action Learning  & 
    Hybrid AR+Diffusion & Img, Act & Img, Act & nuScenes~\cite{nuscenes}  \\
    \bottomrule
  \end{tabular}%
  }
\end{table*}

\subsection{Deep Learning Based Autonomous Driving}

Autonomous driving presents high demands on data diversity, realism, and real-time decision-making.  Collecting and annotating large-scale driving datasets is costly, and models trained in limited or safe conditions often fail to generalize to rare or hazardous scenarios.  While traditional simulators provide some relief, they lack the visual and behavioral richness of real-world traffic.

Interactive Generative Video (IGV) models address these challenges by learning to simulate driving scenes directly from data.  IGV serves both as a realistic video-based simulator for generating diverse training scenarios, and as an internal world model that predicts future outcomes conditioned on control inputs.  These capabilities support two key directions which shown in the Tab.~\ref{tab:app_ad} and Fig.~\ref{fig:ad}: (1) large-scale video simulation for offline training and evaluation, and (2) video-based future reasoning for real-time planning and control.
 
\subsection{Video Simulation for Autonomous Driving}

For control problems where unlimited interaction with the real world is impractical, simulators are typically either hand-coded with domain-specific rules or built from data.  However, traditional simulators—based on fixed physics engines and stochastic models—struggle to capture the full complexity and variability of real-world driving.  In contrast, IGV offers a data-driven alternative by simulating realistic video streams conditioned on interactive control signals, serving as a learned visual world model.

By leveraging the Generation and Dynamics modules, IGV can synthesize high-fidelity video sequences that evolve in response to driver actions such as steering or acceleration, accurately modeling how the environment changes under physical constraints.  The Control module ensures that these simulations can be interactively steered in real-time, reflecting fine-grained agent inputs.  Meanwhile, the Memory module maintains continuity in visual and spatial context—such as consistent object layouts and traffic flows—across simulated frames.  Together, these modules allow IGV to simulate diverse driving trajectories and environmental responses, offering a flexible, realistic platform for training, testing, and planning in autonomous driving systems.

DriveSim~\cite{drivesim} employs action-conditioned Recurrent Neural Networks to learn a transition model within the embedded space, enabling the generation of realistic-looking video predictions for multiple consecutive frames. 
DriveGAN~\cite{drivegan} leverages Generative Adversarial Networks (GANs) to model and simulate dynamic environments directly in pixel space. By disentangling scene components without supervision, the GAN-based framework enables realistic and controllable video generation, offering flexibility in manipulating elements such as weather, object locations, and driving controls.
GAIA-1~\cite{gaia-1} is a diffusion based generative world model for autonomous driving that uses video, text, and action inputs to simulate realistic driving scenarios, offering fine-grained control over vehicle behavior and scene features, while enabling advanced training and innovation in autonomous systems.
DriveDreamer~\cite{drivedreamer} utilize a diffusion model and a two-stage training pipeline to capture structured traffic constraints and predict future states, enabling precise, controllable video generation.
WoVoGen~\cite{wovogen} is a novel multi-camera driving scene generator that leverages a 4D world volume to ensure intra-world consistency and inter-sensor coherence in diffusion-based video generation.
Panacea~\cite{wen2024panacea} and DrivingDiffusion~\cite{drivingdiffusion} address the need for high-quality, large-scale annotated multi-view driving videos by ensuring cross-view and temporal consistency while providing precise controllability. Together, they enable realistic video generation controlled by BEV layouts, significantly enhancing datasets for autonomous driving research and development.
GenAD~\cite{genad} and Vista~\cite{vista} pre-train on high-definition, unlabeled autonomous vehicle video data sourced from the internet, and fine-tune using small amounts of video data with multiple control annotations to enhance consistency and improve signal control.

Together, these systems showcase IGV’s strength in simulating realistic, controllable, and diverse driving scenarios for safe and efficient autonomous driving development.

\subsection{Driving with Generative Videos}

In autonomous driving, decision-making under real-time constraints requires the system to anticipate and evaluate potential outcomes of its actions. Traditional methods often separate planning and simulation, relying on static models or hand-coded rules to simulate driving scenarios. However, these approaches lack the flexibility and realism needed to handle dynamic environments. IGV offers a more robust solution by integrating video prediction with control decision-making, allowing for real-time reasoning about future scenarios.

By leveraging the Generation and Dynamics modules, IGV can predict high-fidelity visual sequences based on control signals such as steering, speed, or braking, capturing the evolving traffic and environmental conditions in response to driver actions. The Control module allows the vehicle to steer these simulations interactively, adjusting for real-time changes in environment or behavior. Meanwhile, the Memory module ensures spatial and temporal continuity, preserving the layout of dynamic objects like vehicles and pedestrians across generated frames. Together, these modules enable IGV to provide an intuitive, interpretable decision-making framework where the vehicle can visualize potential outcomes of its actions before executing them, enhancing planning safety and decision quality.

Drive-WM~\cite{Drive-WM} is a novel driving world model that generates high-fidelity multiview videos to simulate multiple future scenarios based on distinct driving maneuvers, enabling safe and efficient planning through image-based rewards.
ADriver-I~\cite{adriver} is a unified world model for autonomous driving that leverages multimodal large language models and diffusion techniques to process interleaved vision-action pairs, autoregressively predicting control signals and future frames.

These approaches enhance interpretability, as developers can visualize predicted outcomes and understand the reasoning behind decisions. IGV-based models like these allow for more flexible, interpretable, and robust autonomous driving policies.

\subsection{Challenges and Future Work}

Interactive Generative Videos (IGV) hold great promise for advancing autonomous driving by offering realistic simulation and real-time planning capabilities. However, deploying these techniques in safety-critical driving systems presents several challenges. Below, we outline two key challenges and suggest directions for future research.

\subsubsection{Challenge \#1: Enhancing Realism and Generalization in IGV-Based Driving Simulators}
\textbf{Related to:} \textit{Video Simulation for Autonomous Driving}

While IGV-based simulators can generate high-fidelity driving videos from data, they must accurately capture the complexity and variability of real-world traffic. Traditional simulators rely on fixed physics engines and stochastic models, but these often fail to reflect subtle physical dynamics and diverse environmental conditions. IGV, despite its data-driven advantages, still faces the challenge of generating visually and behaviorally authentic scenarios that generalize across different road conditions, lighting, weather, and rare events.

\textbf{Future direction:} Future research should focus on integrating advanced physical models and multi-modal sensor fusion techniques into the IGV framework to enhance simulation realism. Techniques such as domain randomization, real-world fine-tuning, and curriculum learning could help broaden the training distribution, while innovative methods for incorporating temporal consistency and cross-view information will further improve the generalization of IGV-based simulators.

\subsubsection{Challenge \#2: Achieving Real-Time Interactivity and System Integration for IGV-Based Driving}
\textbf{Related to:} \textit{Driving with Interactive Generative Videos}

For real-time decision-making, IGV must not only predict high-fidelity future scenes but also integrate seamlessly with control systems to adapt on the fly. Existing approaches often decouple planning from simulation, and even when IGV is used, achieving low-latency interactive control that reflects subtle driver inputs remains challenging. Moreover, merging IGV-generated scenarios with actual vehicle control architectures requires robust handling of data flow and precise alignment between simulated predictions and real-world sensor inputs.

\textbf{Future direction:} Research should aim to reduce latency in video prediction and improve the responsiveness of IGV-based control loops by optimizing generative architectures and enhancing the real-time performance of the Control module. Additionally, developing integration protocols and adaptive feedback mechanisms that align IGV outputs with physical control systems will be crucial for ensuring that the visual predictions reliably inform safe, real-world driving actions.




%% file: IGV_sections/7_conclusion.tex
\section{Conclusion}

This paper presents a comprehensive survey of Interactive Generative Video (IGV) technology through three significant contributions. First, we conducted an extensive survey of IGV applications across gaming, embodied AI, and autonomous driving domains, providing a thorough overview of current technological capabilities. Second, we introduced a systematic framework decomposing IGV systems into five essential modules: Generation, Control, Memory, Dynamics, and Intelligence, offering a structured approach to understanding and developing IGV systems. Third, through this framework, we systematically analyzed the technical challenges in realizing each component, identifying critical areas requiring further research. Our work serves as a foundational resource for future IGV research by providing a clear roadmap of current capabilities and development directions, which we believe will facilitate more focused research efforts toward advancing sophisticated and practical applications across various domains.

%% file: main.bbl
\begin{thebibliography}{100}
\providecommand{\url}[1]{#1}
\csname url@samestyle\endcsname
\providecommand{\newblock}{\relax}
\providecommand{\bibinfo}[2]{#2}
\providecommand{\BIBentrySTDinterwordspacing}{\spaceskip=0pt\relax}
\providecommand{\BIBentryALTinterwordstretchfactor}{4}
\providecommand{\BIBentryALTinterwordspacing}{\spaceskip=\fontdimen2\font plus
\BIBentryALTinterwordstretchfactor\fontdimen3\font minus \fontdimen4\font\relax}
\providecommand{\BIBforeignlanguage}[2]{{%
\expandafter\ifx\csname l@#1\endcsname\relax
\typeout{** WARNING: IEEEtran.bst: No hyphenation pattern has been}%
\typeout{** loaded for the language `#1'. Using the pattern for}%
\typeout{** the default language instead.}%
\else
\language=\csname l@#1\endcsname
\fi
#2}}
\providecommand{\BIBdecl}{\relax}
\BIBdecl

\bibitem{sora}
OpenAI, ``Creating video from text,'' \url{https://openai.com/index/sora/}, 2024.

\bibitem{cogvideox}
Z.~Yang, J.~Teng, W.~Zheng, M.~Ding, S.~Huang, J.~Xu, Y.~Yang, W.~Hong, X.~Zhang, G.~Feng \emph{et~al.}, ``Cogvideox: Text-to-video diffusion models with an expert transformer,'' \emph{arXiv preprint arXiv:2408.06072}, 2024.

\bibitem{kling}
Kling, ``Kling ai: Next-generation ai creative studio,'' \url{https://app.klingai.com/}, 2024.

\bibitem{veo2}
G.~DeepMind, ``Veo 2: Our state-of-the-art video generation model,'' \url{https://deepmind.google/technologies/veo/veo-2/}, 2024.

\bibitem{seaweed}
T.~Seawead, C.~Yang, Z.~Lin, Y.~Zhao, S.~Lin, Z.~Ma, H.~Guo, H.~Chen, L.~Qi, S.~Wang \emph{et~al.}, ``Seaweed-7b: Cost-effective training of video generation foundation model,'' \emph{arXiv preprint arXiv:2504.08685}, 2025.

\bibitem{moviegen}
T.~M.~G. team, ``Movie gen: A cast of media foundation models,'' \emph{arXiv preprint arXiv:2410.13720}, 2024.

\bibitem{runway}
Runway, ``Runway : Tools for human imagination.'' \url{https://runwayml.com/}, 2024.

\bibitem{luma}
L.~AI, ``Luma ai,'' \url{https://lumalabs.ai/}, 2024.

\bibitem{vidu}
F.~Bao, C.~Xiang, G.~Yue, G.~He, H.~Zhu, K.~Zheng, M.~Zhao, S.~Liu, Y.~Wang, and J.~Zhu, ``Vidu: a highly consistent, dynamic and skilled text-to-video generator with diffusion models,'' \emph{arXiv preprint arXiv:2405.04233}, 2024.

\bibitem{wan}
A.~Wang, B.~Ai, B.~Wen, C.~Mao, C.-W. Xie, D.~Chen, F.~Yu, H.~Zhao, J.~Yang, J.~Zeng \emph{et~al.}, ``Wan: Open and advanced large-scale video generative models,'' \emph{arXiv preprint arXiv:2503.20314}, 2025.

\bibitem{hunyuanvideo}
W.~Kong, Q.~Tian, Z.~Zhang, R.~Min, Z.~Dai, J.~Zhou, J.~Xiong, X.~Li, B.~Wu, J.~Zhang \emph{et~al.}, ``Hunyuanvideo: A systematic framework for large video generative models,'' \emph{arXiv preprint arXiv:2412.03603}, 2024.

\bibitem{ddpm}
J.~Ho, A.~Jain, and P.~Abbeel, ``Denoising diffusion probabilistic models,'' \emph{Advances in neural information processing systems}, 2020.

\bibitem{score}
Y.~Song and S.~Ermon, ``Generative modeling by estimating gradients of the data distribution,'' \emph{Advances in neural information processing systems}, 2019.

\bibitem{flow}
Y.~Lipman, R.~T. Chen, H.~Ben-Hamu, M.~Nickel, and M.~Le, ``Flow matching for generative modeling,'' \emph{arXiv preprint arXiv:2210.02747}, 2022.

\bibitem{rectified}
X.~Liu, C.~Gong, and Q.~Liu, ``Flow straight and fast: Learning to generate and transfer data with rectified flow,'' \emph{arXiv preprint arXiv:2209.03003}, 2022.

\bibitem{videogpt}
W.~Yan, Y.~Zhang, P.~Abbeel, and A.~Srinivas, ``Videogpt: Video generation using vq-vae and transformers,'' \emph{arXiv preprint arXiv:2104.10157}, 2021.

\bibitem{videopoet}
D.~Kondratyuk, L.~Yu, X.~Gu, J.~Lezama, J.~Huang, G.~Schindler, R.~Hornung, V.~Birodkar, J.~Yan, M.-C. Chiu \emph{et~al.}, ``Videopoet: A large language model for zero-shot video generation,'' \emph{arXiv preprint arXiv:2312.14125}, 2023.

\bibitem{emu3}
X.~Wang, X.~Zhang, Z.~Luo, Q.~Sun, Y.~Cui, J.~Wang, F.~Zhang, Y.~Wang, Z.~Li, Q.~Yu \emph{et~al.}, ``Emu3: Next-token prediction is all you need,'' \emph{arXiv preprint arXiv:2409.18869}, 2024.

\bibitem{gamengine}
D.~Valevski, Y.~Leviathan, M.~Arar, and S.~Fruchter, ``Diffusion models are real-time game engines,'' \emph{arXiv preprint arXiv:2408.14837}, 2024.

\bibitem{gamegen-x}
H.~Che, X.~He, Q.~Liu, C.~Jin, and H.~Chen, ``Gamegen-x: Interactive open-world game video generation,'' \emph{arXiv preprint arXiv:2411.00769}, 2024.

\bibitem{matrix}
R.~Feng, H.~Zhang, Z.~Yang, J.~Xiao, Z.~Shu, Z.~Liu, A.~Zheng, Y.~Huang, Y.~Liu, and H.~Zhang, ``The matrix: Infinite-horizon world generation with real-time moving control,'' \emph{arXiv preprint arXiv:2412.03568}, 2024.

\bibitem{oasis}
E.~Decart, ``Oasis: A universe in a transformer,'' \url{https://oasis-model.github.io/}, 2024.

\bibitem{gamefactory}
J.~Yu, Y.~Qin, X.~Wang, P.~Wan, D.~Zhang, and X.~Liu, ``Gamefactory: Creating new games with generative interactive videos,'' 2025.

\bibitem{genie}
J.~Bruce, M.~D. Dennis, A.~Edwards, J.~Parker-Holder, Y.~Shi, E.~Hughes, M.~Lai, A.~Mavalankar, R.~Steigerwald, C.~Apps \emph{et~al.}, ``Genie: Generative interactive environments,'' in \emph{International Conference on Machine Learning}, 2024.

\bibitem{genie2}
\BIBentryALTinterwordspacing
DeepMind, ``Genie 2: A large-scale foundation world model,'' 2025, accessed: 2025-01-17. [Online]. Available: \url{https://deepmind.google/discover/blog/genie-2-a-large-scale-foundation-world-model/}
\BIBentrySTDinterwordspacing

\bibitem{wham}
A.~Kanervisto, D.~Bignell, L.~Y. Wen, M.~Grayson, R.~Georgescu, S.~Valcarcel~Macua, S.~Z. Tan, T.~Rashid, T.~Pearce, Y.~Cao \emph{et~al.}, ``World and human action models towards gameplay ideation,'' \emph{Nature}, vol. 638, no. 8051, pp. 656--663, 2025.

\bibitem{gamegan}
S.~W. Kim, Y.~Zhou, J.~Philion, A.~Torralba, and S.~Fidler, ``Learning to simulate dynamic environments with gamegan,'' in \emph{Proceedings of the IEEE/CVF Conference on Computer Vision and Pattern Recognition}, 2020, pp. 1231--1240.

\bibitem{caddy}
W.~Menapace, S.~Lathuiliere, S.~Tulyakov, A.~Siarohin, and E.~Ricci, ``Playable video generation,'' in \emph{Proceedings of the IEEE/CVF Conference on Computer Vision and Pattern Recognition}, 2021, pp. 10\,061--10\,070.

\bibitem{pe}
W.~Menapace, S.~Lathuili{\`e}re, A.~Siarohin, C.~Theobalt, S.~Tulyakov, V.~Golyanik, and E.~Ricci, ``Playable environments: Video manipulation in space and time,'' in \emph{Proceedings of the IEEE/CVF Conference on Computer Vision and Pattern Recognition}, 2022, pp. 3584--3593.

\bibitem{pgm}
W.~Menapace, A.~Siarohin, S.~Lathuili{\`e}re, P.~Achlioptas, V.~Golyanik, S.~Tulyakov, and E.~Ricci, ``Promptable game models: Text-guided game simulation via masked diffusion models,'' \emph{ACM Transactions on Graphics}, vol.~43, no.~2, pp. 1--16, 2024.

\bibitem{diamond}
E.~Alonso, A.~Jelley, V.~Micheli, A.~Kanervisto, A.~Storkey, T.~Pearce, and F.~Fleuret, ``Diffusion for world modeling: Visual details matter in atari,'' in \emph{Thirty-eighth Conference on Neural Information Processing Systems}, 2024.

\bibitem{plan4mc}
P.~BAAI, ``Plan4mc: Skill reinforcement learning and planning for open-world minecraft tasks,'' \emph{arXiv preprint arXiv:2303.16563}, 2023.

\bibitem{mineworld}
J.~Guo, Y.~Ye, T.~He, H.~Wu, Y.~Jiang, T.~Pearce, and J.~Bian, ``Mineworld: a real-time and open-source interactive world model on minecraft,'' \emph{arXiv preprint arXiv:2504.08388}, 2025.

\bibitem{maag}
J.~Chen, Y.~Zhao, Y.~Huang, L.~Cui, L.~Dong, T.~Lv, Q.~Chen, and F.~Wei, ``Model as a game: On numerical and spatial consistency for generative games,'' \emph{arXiv preprint arXiv:2503.21172}, 2025.

\bibitem{adaworld}
S.~Gao, S.~Zhou, Y.~Du, J.~Zhang, and C.~Gan, ``Adaworld: Learning adaptable world models with latent actions,'' \emph{arXiv preprint arXiv:2503.18938}, 2025.

\bibitem{vlp}
Y.~Du, M.~Yang, P.~Florence, F.~Xia, A.~Wahid, B.~Ichter, P.~Sermanet, T.~Yu, P.~Abbeel, J.~B. Tenenbaum \emph{et~al.}, ``Video language planning,'' \emph{arXiv preprint arXiv:2310.10625}, 2023.

\bibitem{hip}
A.~Ajay, S.~Han, Y.~Du, S.~Li, A.~Gupta, T.~Jaakkola, J.~Tenenbaum, L.~Kaelbling, A.~Srivastava, and P.~Agrawal, ``Compositional foundation models for hierarchical planning,'' \emph{Advances in Neural Information Processing Systems}, vol.~36, pp. 22\,304--22\,325, 2023.

\bibitem{unipi}
Y.~Du, S.~Yang, B.~Dai, H.~Dai, O.~Nachum, J.~Tenenbaum, D.~Schuurmans, and P.~Abbeel, ``Learning universal policies via text-guided video generation,'' \emph{Advances in neural information processing systems}, vol.~36, pp. 9156--9172, 2023.

\bibitem{unisim}
M.~Yang, Y.~Du, K.~Ghasemipour, J.~Tompson, D.~Schuurmans, and P.~Abbeel, ``Learning interactive real-world simulators,'' \emph{arXiv preprint arXiv:2310.06114}, 2023.

\bibitem{irasim}
F.~Zhu, H.~Wu, S.~Guo, Y.~Liu, C.~Cheang, and T.~Kong, ``Irasim: Learning interactive real-robot action simulators,'' \emph{arXiv preprint arXiv:2406.14540}, 2024.

\bibitem{robodreamer}
S.~Zhou, Y.~Du, J.~Chen, Y.~Li, D.-Y. Yeung, and C.~Gan, ``Robodreamer: Learning compositional world models for robot imagination,'' \emph{arXiv preprint arXiv:2404.12377}, 2024.

\bibitem{combo}
H.~Zhang, Z.~Wang, Q.~Lyu, Z.~Zhang, S.~Chen, T.~Shu, B.~Dariush, K.~Lee, Y.~Du, and C.~Gan, ``Combo: Compositional world models for embodied multi-agent cooperation,'' \emph{arXiv preprint arXiv:2404.10775}, 2024.

\bibitem{worldsimbench}
Y.~Qin, Z.~Shi, J.~Yu, X.~Wang, E.~Zhou, L.~Li, Z.~Yin, X.~Liu, L.~Sheng, J.~Shao \emph{et~al.}, ``Worldsimbench: Towards video generation models as world simulators,'' \emph{arXiv preprint arXiv:2410.18072}, 2024.

\bibitem{4dworldmodel}
H.~Zhen, Q.~Sun, P.~Han, S.~Zhou, Y.~Du, and C.~Gan, ``Learning 4d embodied world models.''

\bibitem{navigatediff}
Y.~Qin, A.~Sun, Y.~Hong, B.~Wang, and R.~Zhang, ``Navigatediff: Visual predictors are zero-shot navigation assistants,'' \emph{arXiv preprint arXiv:2502.13894}, 2025.

\bibitem{gce}
Y.~Luo and Y.~Du, ``Grounding video models to actions through goal conditioned exploration,'' \emph{arXiv preprint arXiv:2411.07223}, 2024.

\bibitem{cosmos}
NVIDIA, ``Cosmos world foundation model platform for physical ai,'' \emph{arXiv preprint arXiv:2501.03575}, 2025.

\bibitem{UVA}
S.~Li, Y.~Gao, D.~Sadigh, and S.~Song, ``Unified video action model,'' \emph{arXiv preprint arXiv:2503.00200}, 2025.

\bibitem{drivegan}
S.~W. Kim, J.~Philion, A.~Torralba, and S.~Fidler, ``Drivegan: Towards a controllable high-quality neural simulation,'' in \emph{Proceedings of IEEE/CVF Conference on Computer Vision and Pattern Recognition}, 2021, pp. 5820--5829.

\bibitem{drivesim}
E.~Santana and G.~Hotz, ``Learning a driving simulator,'' \emph{arXiv preprint arXiv:1608.01230}, 2016.

\bibitem{gaia-1}
A.~Hu, L.~Russell, H.~Yeo, Z.~Murez, G.~Fedoseev, A.~Kendall, J.~Shotton, and G.~Corrado, ``Gaia-1: A generative world model for autonomous driving,'' \emph{arXiv preprint arXiv:2309.17080}, 2023.

\bibitem{wovogen}
J.~Lu, Z.~Huang, Z.~Yang, J.~Zhang, and L.~Zhang, ``Wovogen: World volume-aware diffusion for controllable multi-camera driving scene generation,'' in \emph{European Conference on Computer Vision}.\hskip 1em plus 0.5em minus 0.4em\relax Springer, 2024, pp. 329--345.

\bibitem{adriver}
F.~Jia, W.~Mao, Y.~Liu, Y.~Zhao, Y.~Wen, C.~Zhang, X.~Zhang, and T.~Wang, ``Adriver-i: A general world model for autonomous driving,'' \emph{arXiv preprint arXiv:2311.13549}, 2023.

\bibitem{wen2024panacea}
Y.~Wen, Y.~Zhao, Y.~Liu, F.~Jia, Y.~Wang, C.~Luo, C.~Zhang, T.~Wang, X.~Sun, and X.~Zhang, ``Panacea: Panoramic and controllable video generation for autonomous driving,'' in \emph{Proceedings of the IEEE/CVF Conference on Computer Vision and Pattern Recognition}, 2024, pp. 6902--6912.

\bibitem{Drive-WM}
Y.~Wang, J.~He, L.~Fan, H.~Li, Y.~Chen, and Z.~Zhang, ``Driving into the future: Multiview visual forecasting and planning with world model for autonomous driving,'' in \emph{Proceedings of the IEEE/CVF Conference on Computer Vision and Pattern Recognition}, 2024, pp. 14\,749--14\,759.

\bibitem{drivingdiffusion}
X.~Li, Y.~Zhang, and X.~Ye, ``Drivingdiffusion: Layout-guided multi-view driving scenarios video generation with latent diffusion model,'' in \emph{European Conference on Computer Vision}.\hskip 1em plus 0.5em minus 0.4em\relax Springer, 2024, pp. 469--485.

\bibitem{drivedreamer}
X.~Wang, Z.~Zhu, G.~Huang, X.~Chen, J.~Zhu, and J.~Lu, ``Drivedreamer: Towards real-world-drive world models for autonomous driving,'' in \emph{European Conference on Computer Vision}.\hskip 1em plus 0.5em minus 0.4em\relax Springer, 2024, pp. 55--72.

\bibitem{drivedreamer-2}
G.~Zhao, X.~Wang, Z.~Zhu, X.~Chen, G.~Huang, X.~Bao, and X.~Wang, ``Drivedreamer-2: Llm-enhanced world models for diverse driving video generation,'' in \emph{Proceedings of the AAAI Conference on Artificial Intelligence}, vol.~39, no.~10, 2025, pp. 10\,412--10\,420.

\bibitem{genad}
W.~Zheng, R.~Song, X.~Guo, C.~Zhang, and L.~Chen, ``Genad: Generative end-to-end autonomous driving,'' in \emph{European Conference on Computer Vision}.\hskip 1em plus 0.5em minus 0.4em\relax Springer, 2024, pp. 87--104.

\bibitem{vista}
S.~Gao, J.~Yang, L.~Chen, K.~Chitta, Y.~Qiu, A.~Geiger, J.~Zhang, and H.~Li, ``Vista: A generalizable driving world model with high fidelity and versatile controllability,'' \emph{arXiv preprint arXiv:2405.17398}, 2024.

\bibitem{gaia-2}
L.~Russell, A.~Hu, L.~Bertoni, G.~Fedoseev, J.~Shotton, E.~Arani, and G.~Corrado, ``Gaia-2: A controllable multi-view generative world model for autonomous driving,'' \emph{arXiv preprint arXiv:2503.20523}, 2025.

\bibitem{cogen}
Y.~Ji, Z.~Zhu, Z.~Zhu, K.~Xiong, M.~Lu, Z.~Li, L.~Zhou, H.~Sun, B.~Wang, and T.~Lu, ``Cogen: 3d consistent video generation via adaptive conditioning for autonomous driving,'' \emph{arXiv preprint arXiv:2503.22231}, 2025.

\bibitem{drivedreamer4d}
G.~Zhao, C.~Ni, X.~Wang, Z.~Zhu, X.~Zhang, Y.~Wang, G.~Huang, X.~Chen, B.~Wang, Y.~Zhang \emph{et~al.}, ``Drivedreamer4d: World models are effective data machines for 4d driving scene representation,'' \emph{arXiv preprint arXiv:2410.13571}, 2024.

\bibitem{maskgwm}
J.~Ni, Y.~Guo, Y.~Liu, R.~Chen, L.~Lu, and Z.~Wu, ``Maskgwm: A generalizable driving world model with video mask reconstruction,'' \emph{arXiv preprint arXiv:2502.11663}, 2025.

\bibitem{igvposition}
J.~Yu, Y.~Qin, H.~Che, Q.~Liu, X.~Wang, P.~Wan, D.~Zhang, and X.~Liu, ``Position: Interactive generative video as next-generation game engine,'' \emph{arXiv preprint arXiv:2503.17359}, 2025.

\bibitem{kingma2019introduction}
D.~P. Kingma, M.~Welling \emph{et~al.}, ``An introduction to variational autoencoders,'' \emph{Foundations and Trends{\textregistered} in Machine Learning}, vol.~12, no.~4, pp. 307--392, 2019.

\bibitem{GANs}
A.~Creswell, T.~White, V.~Dumoulin, K.~Arulkumaran, B.~Sengupta, and A.~A. Bharath, ``Generative adversarial networks: An overview,'' \emph{IEEE signal processing magazine}, vol.~35, no.~1, pp. 53--65, 2018.

\bibitem{ho2020denoising}
J.~Ho, A.~Jain, and P.~Abbeel, ``Denoising diffusion probabilistic models,'' \emph{Advances in neural information processing systems}, vol.~33, pp. 6840--6851, 2020.

\bibitem{reflow}
X.~Liu, C.~Gong, and Q.~Liu, ``Flow straight and fast: Learning to generate and transfer data with rectified flow,'' \emph{arXiv preprint arXiv:2209.03003}, 2022.

\bibitem{flowmatching}
Y.~Lipman, R.~T. Chen, H.~Ben-Hamu, M.~Nickel, and M.~Le, ``Flow matching for generative modeling,'' \emph{arXiv preprint arXiv:2210.02747}, 2022.

\bibitem{xiong2024autoregressive}
J.~Xiong, G.~Liu, L.~Huang, C.~Wu, T.~Wu, Y.~Mu, Y.~Yao, H.~Shen, Z.~Wan, J.~Huang \emph{et~al.}, ``Autoregressive models in vision: A survey,'' \emph{arXiv preprint arXiv:2411.05902}, 2024.

\bibitem{diffforcing}
B.~Chen, D.~M. Monso, Y.~Du, M.~Simchowitz, R.~Tedrake, and V.~Sitzmann, ``Diffusion forcing: Next-token prediction meets full-sequence diffusion,'' \emph{arXiv preprint arXiv:2407.01392}, 2024.

\bibitem{causaldiff}
C.~Deng, D.~Zh, K.~Li, S.~Guan, and H.~Fan, ``Causal diffusion transformers for generative modeling,'' \emph{arXiv preprint arXiv:2412.12095}, 2024.

\bibitem{tulyakov2018mocogan}
S.~Tulyakov, M.-Y. Liu, X.~Yang, and J.~Kautz, ``Mocogan: Decomposing motion and content for video generation,'' in \emph{Proceedings of the IEEE conference on computer vision and pattern recognition}, 2018, pp. 1526--1535.

\bibitem{ho2022video}
J.~Ho, T.~Salimans, A.~Gritsenko, W.~Chan, M.~Norouzi, and D.~J. Fleet, ``Video diffusion models,'' \emph{Advances in Neural Information Processing Systems}, vol.~35, pp. 8633--8646, 2022.

\bibitem{blattmann2023stable}
A.~Blattmann, T.~Dockhorn, S.~Kulal, D.~Mendelevitch, M.~Kilian, D.~Lorenz, Y.~Levi, Z.~English, V.~Voleti, A.~Letts \emph{et~al.}, ``Stable video diffusion: Scaling latent video diffusion models to large datasets,'' \emph{arXiv preprint arXiv:2311.15127}, 2023.

\bibitem{harvey2022flexible}
W.~Harvey, S.~Naderiparizi, V.~Masrani, C.~Weilbach, and F.~Wood, ``Flexible diffusion modeling of long videos,'' \emph{Advances in Neural Information Processing Systems}, vol.~35, pp. 27\,953--27\,965, 2022.

\bibitem{luo2023videofusion}
Z.~Luo, D.~Chen, Y.~Zhang, Y.~Huang, L.~Wang, Y.~Shen, D.~Zhao, J.~Zhou, and T.~Tan, ``Videofusion: Decomposed diffusion models for high-quality video generation,'' \emph{arXiv preprint arXiv:2303.08320}, 2023.

\bibitem{chen2023videocrafter1}
H.~Chen, M.~Xia, Y.~He, Y.~Zhang, X.~Cun, S.~Yang, J.~Xing, Y.~Liu, Q.~Chen, X.~Wang \emph{et~al.}, ``Videocrafter1: Open diffusion models for high-quality video generation,'' \emph{arXiv preprint arXiv:2310.19512}, 2023.

\bibitem{chen2024videocrafter2}
H.~Chen, Y.~Zhang, X.~Cun, M.~Xia, X.~Wang, C.~Weng, and Y.~Shan, ``Videocrafter2: Overcoming data limitations for high-quality video diffusion models,'' in \emph{Proceedings of the IEEE/CVF Conference on Computer Vision and Pattern Recognition}, 2024, pp. 7310--7320.

\bibitem{he2022latent}
Y.~He, T.~Yang, Y.~Zhang, Y.~Shan, and Q.~Chen, ``Latent video diffusion models for high-fidelity long video generation,'' \emph{arXiv preprint arXiv:2211.13221}, 2022.

\bibitem{opensora}
Z.~Zheng, X.~Peng, T.~Yang, C.~Shen, S.~Li, H.~Liu, Y.~Zhou, T.~Li, and Y.~You, ``Open-sora: Democratizing efficient video production for all,'' \url{https://github.com/hpcaitech/Open-Sora}, 2024.

\bibitem{kong2024hunyuanvideo}
\BIBentryALTinterwordspacing
W.~Kong, Q.~Tian, Z.~Zhang, R.~Min, Z.~Dai, J.~Zhou, J.~Xiong, X.~Li, B.~Wu, J.~Zhang, K.~Wu, Q.~Lin, A.~Wang, A.~Wang, C.~Li, D.~Huang, F.~Yang, H.~Tan, H.~Wang, J.~Song, J.~Bai, J.~Wu, J.~Xue, J.~Wang, J.~Yuan, K.~Wang, M.~Liu, P.~Li, S.~Li, W.~Wang, W.~Yu, X.~Deng, Y.~Li, Y.~Long, Y.~Chen, Y.~Cui, Y.~Peng, Z.~Yu, Z.~He, Z.~Xu, Z.~Zhou, Z.~Xu, Y.~Tao, Q.~Lu, S.~Liu, D.~Zhou, H.~Wang, Y.~Yang, D.~Wang, Y.~Liu, and J.~Jiang, ``Hunyuanvideo: A systematic framework for large video generative models,'' 2024, along with contributions from Caesar Zhong. [Online]. Available: \url{https://arxiv.org/abs/2412.03603}
\BIBentrySTDinterwordspacing

\bibitem{chen2025goku}
S.~Chen, C.~Ge, Y.~Zhang, Y.~Zhang, F.~Zhu, H.~Yang, H.~Hao, H.~Wu, Z.~Lai, Y.~Hu \emph{et~al.}, ``Goku: Flow based video generative foundation models,'' \emph{arXiv preprint arXiv:2502.04896}, 2025.

\bibitem{chen2020generative}
M.~Chen, A.~Radford, R.~Child, J.~Wu, H.~Jun, D.~Luan, and I.~Sutskever, ``Generative pretraining from pixels,'' in \emph{International conference on machine learning}.\hskip 1em plus 0.5em minus 0.4em\relax PMLR, 2020, pp. 1691--1703.

\bibitem{yu2022scaling}
J.~Yu, Y.~Xu, J.~Y. Koh, T.~Luong, G.~Baid, Z.~Wang, V.~Vasudevan, A.~Ku, Y.~Yang, B.~K. Ayan \emph{et~al.}, ``Scaling autoregressive models for content-rich text-to-image generation,'' \emph{arXiv preprint arXiv:2206.10789}, vol.~2, no.~3, p.~5, 2022.

\bibitem{team2024chameleon}
C.~Team, ``Chameleon: Mixed-modal early-fusion foundation models,'' \emph{arXiv preprint arXiv:2405.09818}, 2024.

\bibitem{hong2022cogvideo}
W.~Hong, M.~Ding, W.~Zheng, X.~Liu, and J.~Tang, ``Cogvideo: Large-scale pretraining for text-to-video generation via transformers,'' \emph{arXiv preprint arXiv:2205.15868}, 2022.

\bibitem{kondratyuk2023videopoet}
D.~Kondratyuk, L.~Yu, X.~Gu, J.~Lezama, J.~Huang, G.~Schindler, R.~Hornung, V.~Birodkar, J.~Yan, M.-C. Chiu \emph{et~al.}, ``Videopoet: A large language model for zero-shot video generation,'' \emph{arXiv preprint arXiv:2312.14125}, 2023.

\bibitem{wang2024emu3}
X.~Wang, X.~Zhang, Z.~Luo, Q.~Sun, Y.~Cui, J.~Wang, F.~Zhang, Y.~Wang, Z.~Li, Q.~Yu \emph{et~al.}, ``Emu3: Next-token prediction is all you need,'' \emph{arXiv preprint arXiv:2409.18869}, 2024.

\bibitem{wang2024loong}
Y.~Wang, T.~Xiong, D.~Zhou, Z.~Lin, Y.~Zhao, B.~Kang, J.~Feng, and X.~Liu, ``Loong: Generating minute-level long videos with autoregressive language models,'' \emph{arXiv preprint arXiv:2410.02757}, 2024.

\bibitem{yan2021videogpt}
W.~Yan, Y.~Zhang, P.~Abbeel, and A.~Srinivas, ``Videogpt: Video generation using vq-vae and transformers,'' \emph{arXiv preprint arXiv:2104.10157}, 2021.

\bibitem{wu2021godiva}
C.~Wu, L.~Huang, Q.~Zhang, B.~Li, L.~Ji, F.~Yang, G.~Sapiro, and N.~Duan, ``Godiva: Generating open-domain videos from natural descriptions,'' \emph{arXiv preprint arXiv:2104.14806}, 2021.

\bibitem{ge2022long}
S.~Ge, T.~Hayes, H.~Yang, X.~Yin, G.~Pang, D.~Jacobs, J.-B. Huang, and D.~Parikh, ``Long video generation with time-agnostic vqgan and time-sensitive transformer,'' in \emph{European Conference on Computer Vision}.\hskip 1em plus 0.5em minus 0.4em\relax Springer, 2022, pp. 102--118.

\bibitem{sun2025ar}
M.~Sun, W.~Wang, G.~Li, J.~Liu, J.~Sun, W.~Feng, S.~Lao, S.~Zhou, Q.~He, and J.~Liu, ``Ar-diffusion: Asynchronous video generation with auto-regressive diffusion,'' \emph{arXiv preprint arXiv:2503.07418}, 2025.

\bibitem{song2025history}
K.~Song, B.~Chen, M.~Simchowitz, Y.~Du, R.~Tedrake, and V.~Sitzmann, ``History-guided video diffusion,'' \emph{arXiv preprint arXiv:2502.06764}, 2025.

\bibitem{yin2024slow}
T.~Yin, Q.~Zhang, R.~Zhang, W.~T. Freeman, F.~Durand, E.~Shechtman, and X.~Huang, ``From slow bidirectional to fast autoregressive video diffusion models,'' \emph{arXiv preprint arXiv:2412.07772}, vol.~2, 2024.

\bibitem{emu2}
Q.~Sun, Y.~Cui, X.~Zhang, F.~Zhang, Q.~Yu, Y.~Wang, Y.~Rao, J.~Liu, T.~Huang, and X.~Wang, ``Generative multimodal models are in-context learners,'' in \emph{Proceedings of the IEEE/CVF Conference on Computer Vision and Pattern Recognition}, 2024, pp. 14\,398--14\,409.

\bibitem{ge2024seed}
Y.~Ge, S.~Zhao, J.~Zhu, Y.~Ge, K.~Yi, L.~Song, C.~Li, X.~Ding, and Y.~Shan, ``Seed-x: Multimodal models with unified multi-granularity comprehension and generation,'' \emph{arXiv preprint arXiv:2404.14396}, 2024.

\bibitem{zhou2024transfusion}
C.~Zhou, L.~Yu, A.~Babu, K.~Tirumala, M.~Yasunaga, L.~Shamis, J.~Kahn, X.~Ma, L.~Zettlemoyer, and O.~Levy, ``Transfusion: Predict the next token and diffuse images with one multi-modal model,'' \emph{arXiv preprint arXiv:2408.11039}, 2024.

\bibitem{xie2024show}
J.~Xie, W.~Mao, Z.~Bai, D.~J. Zhang, W.~Wang, K.~Q. Lin, Y.~Gu, Z.~Chen, Z.~Yang, and M.~Z. Shou, ``Show-o: One single transformer to unify multimodal understanding and generation,'' \emph{arXiv preprint arXiv:2408.12528}, 2024.

\bibitem{mar}
T.~Li, Y.~Tian, H.~Li, M.~Deng, and K.~He, ``Autoregressive image generation without vector quantization,'' \emph{arXiv preprint arXiv:2406.11838}, 2024.

\bibitem{deng2024autoregressive}
H.~Deng, T.~Pan, H.~Diao, Z.~Luo, Y.~Cui, H.~Lu, S.~Shan, Y.~Qi, and X.~Wang, ``Autoregressive video generation without vector quantization,'' \emph{arXiv preprint arXiv:2412.14169}, 2024.

\bibitem{causvid}
T.~Yin, Q.~Zhang, R.~Zhang, W.~T. Freeman, F.~Durand, E.~Shechtman, and X.~Huang, ``From slow bidirectional to fast causal video generators,'' \emph{arXiv preprint arXiv:2412.07772}, 2024.

\bibitem{rolling}
D.~Ruhe, J.~Heek, T.~Salimans, and E.~Hoogeboom, ``Rolling diffusion models,'' \emph{arXiv preprint arXiv:2402.09470}, 2024.

\bibitem{thematrix}
R.~Feng, H.~Zhang, Z.~Yang, J.~Xiao, Z.~Shu, Z.~Liu, A.~Zheng, Y.~Huang, Y.~Liu, and H.~Zhang, ``The matrix: Infinite-horizon world generation with real-time moving control,'' \emph{arXiv preprint arXiv:2412.03568}, 2024.

\bibitem{oasis2024}
\BIBentryALTinterwordspacing
Decart, J.~Quevedo, Q.~McIntyre, S.~Campbell, X.~Chen, and R.~Wachen, ``Oasis: A universe in a transformer,'' 2024. [Online]. Available: \url{https://oasis-model.github.io/}
\BIBentrySTDinterwordspacing

\bibitem{nova}
H.~Deng, T.~Pan, H.~Diao, Z.~Luo, Y.~Cui, H.~Lu, S.~Shan, Y.~Qi, and X.~Wang, ``Autoregressive video generation without vector quantization,'' \emph{arXiv preprint arXiv:2412.14169}, 2024.

\bibitem{bk-sdm}
B.-K. Kim, H.-K. Song, T.~Castells, and S.~Choi, ``Bk-sdm: A lightweight, fast, and cheap version of stable diffusion,'' in \emph{European Conference on Computer Vision}.\hskip 1em plus 0.5em minus 0.4em\relax Springer, 2025, pp. 381--399.

\bibitem{videolcm}
X.~Wang, S.~Zhang, H.~Zhang, Y.~Liu, Y.~Zhang, C.~Gao, and N.~Sang, ``Videolcm: Video latent consistency model,'' \emph{arXiv preprint arXiv:2312.09109}, 2023.

\bibitem{dcae}
J.~Chen, H.~Cai, J.~Chen, E.~Xie, S.~Yang, H.~Tang, M.~Li, Y.~Lu, and S.~Han, ``Deep compression autoencoder for efficient high-resolution diffusion models,'' \emph{arXiv preprint arXiv:2410.10733}, 2024.

\bibitem{gato}
S.~Reed, K.~Zolna, E.~Parisotto, S.~G. Colmenarejo, A.~Novikov, G.~Barth-Maron, M.~Gimenez, Y.~Sulsky, J.~Kay, J.~T. Springenberg \emph{et~al.}, ``A generalist agent,'' \emph{arXiv preprint arXiv:2205.06175}, 2022.

\bibitem{pengi}
S.~Deshmukh, B.~Elizalde, R.~Singh, and H.~Wang, ``Pengi: An audio language model for audio tasks,'' in \emph{Advances in Neural Information Processing Systems}, 2023.

\bibitem{depthanything}
L.~Yang, B.~Kang, Z.~Huang, X.~Xu, J.~Feng, and H.~Zhao, ``Depth anything: Unleashing the power of large-scale unlabeled data,'' in \emph{Proceedings of the IEEE/CVF Conference on Computer Vision and Pattern Recognition}, 2024, pp. 10\,371--10\,381.

\bibitem{motiongpt}
B.~Jiang, X.~Chen, W.~Liu, J.~Yu, G.~Yu, and T.~Chen, ``Motiongpt: Human motion as a foreign language,'' \emph{Advances in Neural Information Processing Systems}, vol.~36, pp. 20\,067--20\,079, 2023.

\bibitem{lvm}
Y.~Bai, X.~Geng, K.~Mangalam, A.~Bar, A.~L. Yuille, T.~Darrell, J.~Malik, and A.~A. Efros, ``Sequential modeling enables scalable learning for large vision models,'' in \emph{Proceedings of the IEEE/CVF Conference on Computer Vision and Pattern Recognition}, 2024, pp. 22\,861--22\,872.

\bibitem{gamegenx}
H.~Che, X.~He, Q.~Liu, C.~Jin, and H.~Chen, ``Gamegen-x: Interactive open-world game video generation,'' \emph{arXiv preprint arXiv:2411.00769}, 2024.

\bibitem{motionctrl}
Z.~Wang, Z.~Yuan, X.~Wang, Y.~Li, T.~Chen, M.~Xia, P.~Luo, and Y.~Shan, ``Motionctrl: A unified and flexible motion controller for video generation,'' in \emph{ACM SIGGRAPH 2024 Conference Papers}, 2024.

\bibitem{calvin}
O.~Mees, L.~Hermann, E.~Rosete-Beas, and W.~Burgard, ``Calvin: A benchmark for language-conditioned policy learning for long-horizon robot manipulation tasks,'' \emph{IEEE Robotics and Automation Letters}, vol.~7, no.~3, pp. 7327--7334, 2022.

\bibitem{libero}
B.~Liu, Y.~Zhu, C.~Gao, Y.~Feng, Q.~Liu, Y.~Zhu, and P.~Stone, ``Libero: Benchmarking knowledge transfer for lifelong robot learning,'' \emph{Advances in Neural Information Processing Systems}, vol.~36, pp. 44\,776--44\,791, 2023.

\bibitem{robofactory}
Y.~Qin, L.~Kang, X.~Song, Z.~Yin, X.~Liu, X.~Liu, R.~Zhang, and L.~Bai, ``Robofactory: Exploring embodied agent collaboration with compositional constraints,'' \emph{arXiv preprint arXiv:2503.16408}, 2025.

\bibitem{nuscenes}
H.~Caesar, V.~Bankiti, A.~H. Lang, S.~Vora, V.~E. Liong, Q.~Xu, A.~Krishnan, Y.~Pan, G.~Baldan, and O.~Beijbom, ``nuscenes: A multimodal dataset for autonomous driving,'' in \emph{Proceedings of the IEEE/CVF conference on computer vision and pattern recognition}, 2020, pp. 11\,621--11\,631.

\bibitem{waymo}
P.~Sun, H.~Kretzschmar, X.~Dotiwalla, A.~Chouard, V.~Patnaik, P.~Tsui, J.~Guo, Y.~Zhou, Y.~Chai, B.~Caine \emph{et~al.}, ``Scalability in perception for autonomous driving: Waymo open dataset,'' in \emph{Proceedings of the IEEE/CVF conference on computer vision and pattern recognition}, 2020, pp. 2446--2454.

\bibitem{wonderjourney}
H.-X. Yu, H.~Duan, J.~Hur, K.~Sargent, M.~Rubinstein, W.~T. Freeman, F.~Cole, D.~Sun, N.~Snavely, J.~Wu \emph{et~al.}, ``Wonderjourney: Going from anywhere to everywhere,'' in \emph{Proceedings of the IEEE/CVF Conference on Computer Vision and Pattern Recognition}, 2024, pp. 6658--6667.

\bibitem{viewcrafter}
W.~Yu, J.~Xing, L.~Yuan, W.~Hu, X.~Li, Z.~Huang, X.~Gao, T.-T. Wong, Y.~Shan, and Y.~Tian, ``Viewcrafter: Taming video diffusion models for high-fidelity novel view synthesis,'' \emph{arXiv preprint arXiv:2409.02048}, 2024.

\bibitem{see3d}
B.~Ma, H.~Gao, H.~Deng, Z.~Luo, T.~Huang, L.~Tang, and X.~Wang, ``You see it, you got it: Learning 3d creation on pose-free videos at scale,'' \emph{arXiv preprint arXiv:2412.06699}, 2024.

\bibitem{motioncraft}
A.~Montanaro, L.~Savant~Aira, E.~Aiello, D.~Valsesia, and E.~Magli, ``Motioncraft: Physics-based zero-shot video generation,'' \emph{Advances in Neural Information Processing Systems}, vol.~37, pp. 123\,155--123\,181, 2024.

\bibitem{physgen}
S.~Liu, Z.~Ren, S.~Gupta, and S.~Wang, ``Physgen: Rigid-body physics-grounded image-to-video generation,'' in \emph{European Conference on Computer Vision}.\hskip 1em plus 0.5em minus 0.4em\relax Springer, 2024, pp. 360--378.

\bibitem{physdreamer}
T.~Zhang, H.-X. Yu, R.~Wu, B.~Y. Feng, C.~Zheng, N.~Snavely, J.~Wu, and W.~T. Freeman, ``Physdreamer: Physics-based interaction with 3d objects via video generation,'' in \emph{European Conference on Computer Vision}.\hskip 1em plus 0.5em minus 0.4em\relax Springer, 2024, pp. 388--406.

\bibitem{worldmodel}
D.~Ha and J.~Schmidhuber, ``Recurrent world models facilitate policy evolution,'' \emph{Advances in Neural Information Processing Systems}, vol.~31, 2018.

\bibitem{menapace2022playable}
W.~Menapace, S.~Lathuili{\`e}re, A.~Siarohin, C.~Theobalt, S.~Tulyakov, V.~Golyanik, and E.~Ricci, ``Playable environments: Video manipulation in space and time,'' in \emph{Proceedings of the IEEE/CVF Conference on Computer Vision and Pattern Recognition}, 2022, pp. 3584--3593.

\bibitem{menapace2024promptable}
W.~Menapace, A.~Siarohin, S.~Lathuili{\`e}re, P.~Achlioptas, V.~Golyanik, S.~Tulyakov, and E.~Ricci, ``Promptable game models: Text-guided game simulation via masked diffusion models,'' \emph{ACM Transactions on Graphics}, vol.~43, no.~2, pp. 1--16, 2024.

\bibitem{MarioVGG}
\BIBentryALTinterwordspacing
V.~Protocol, ``Video game generation: A practical study using mario,'' 2024, preprint. [Online]. Available: \url{https://github.com/Virtual-Protocol/mario-videogamegen/blob/main/static/pdfs/VideoGameGen.pdf}
\BIBentrySTDinterwordspacing

\bibitem{kazemi2024learning}
N.~Kazemi, N.~Savov, D.~Paudel, and L.~Van~Gool, ``Learning generative interactive environments by trained agent exploration,'' \emph{arXiv preprint arXiv:2409.06445}, 2024.

\bibitem{yang2024playable}
M.~Yang, J.~Li, Z.~Fang, S.~Chen, Y.~Yu, Q.~Fu, W.~Yang, and D.~Ye, ``Playable game generation,'' \emph{arXiv preprint arXiv:2412.00887}, 2024.

\bibitem{bridge}
F.~Ebert, Y.~Yang, K.~Schmeckpeper, B.~Bucher, G.~Georgakis, K.~Daniilidis, C.~Finn, and S.~Levine, ``Bridge data: Boosting generalization of robotic skills with cross-domain datasets,'' \emph{arXiv preprint arXiv:2109.13396}, 2021.

\bibitem{rt-2}
A.~Brohan, N.~Brown, J.~Carbajal, Y.~Chebotar, X.~Chen, K.~Choromanski, T.~Ding, D.~Driess, A.~Dubey, C.~Finn \emph{et~al.}, ``Rt-2: Vision-language-action models transfer web knowledge to robotic control,'' \emph{arXiv preprint arXiv:2307.15818}, 2023.

\bibitem{ego4d}
K.~Grauman, A.~Westbury, E.~Byrne, Z.~Chavis, A.~Furnari, R.~Girdhar, J.~Hamburger, H.~Jiang, M.~Liu, X.~Liu \emph{et~al.}, ``Ego4d: Around the world in 3,000 hours of egocentric video,'' in \emph{Proceedings of the IEEE/CVF conference on computer vision and pattern recognition}, 2022, pp. 18\,995--19\,012.

\bibitem{epic-kitchens}
D.~Damen, H.~Doughty, G.~M. Farinella, S.~Fidler, A.~Furnari, E.~Kazakos, D.~Moltisanti, J.~Munro, T.~Perrett, W.~Price \emph{et~al.}, ``Scaling egocentric vision: The epic-kitchens dataset,'' in \emph{Proceedings of the European conference on computer vision (ECCV)}, 2018, pp. 720--736.

\bibitem{Laion-400m}
C.~Schuhmann, R.~Vencu, R.~Beaumont, R.~Kaczmarczyk, C.~Mullis, A.~Katta, T.~Coombes, J.~Jitsev, and A.~Komatsuzaki, ``Laion-400m: Open dataset of clip-filtered 400 million image-text pairs,'' \emph{arXiv preprint arXiv:2111.02114}, 2021.

\bibitem{genex}
T.~Lu, T.~Shu, J.~Xiao, L.~Ye, J.~Wang, C.~Peng, C.~Wei, D.~Khashabi, R.~Chellappa, A.~Yuille \emph{et~al.}, ``Genex: Generating an explorable world,'' \emph{arXiv preprint arXiv:2412.09624}, 2024.

\bibitem{rt-1}
A.~Brohan, N.~Brown, J.~Carbajal, Y.~Chebotar, J.~Dabis, C.~Finn, K.~Gopalakrishnan, K.~Hausman, A.~Herzog, J.~Hsu \emph{et~al.}, ``Rt-1: Robotics transformer for real-world control at scale,'' \emph{arXiv preprint arXiv:2212.06817}, 2022.

\bibitem{Language-Table}
C.~Lynch, A.~Wahid, J.~Tompson, T.~Ding, J.~Betker, R.~Baruch, T.~Armstrong, and P.~Florence, ``Interactive language: Talking to robots in real time,'' \emph{IEEE Robotics and Automation Letters}, 2023.

\bibitem{habitat}
M.~Savva, A.~Kadian, O.~Maksymets, Y.~Zhao, E.~Wijmans, B.~Jain, J.~Straub, J.~Liu, V.~Koltun, J.~Malik \emph{et~al.}, ``Habitat: A platform for embodied ai research,'' in \emph{Proceedings of the IEEE/CVF international conference on computer vision}, 2019, pp. 9339--9347.

\bibitem{hm3d}
S.~K. Ramakrishnan, A.~Gokaslan, E.~Wijmans, O.~Maksymets, A.~Clegg, J.~Turner, E.~Undersander, W.~Galuba, A.~Westbury, A.~X. Chang \emph{et~al.}, ``Habitat-matterport 3d dataset (hm3d): 1000 large-scale 3d environments for embodied ai,'' \emph{arXiv preprint arXiv:2109.08238}, 2021.

\bibitem{something2something}
R.~Goyal, S.~Ebrahimi~Kahou, V.~Michalski, J.~Materzynska, S.~Westphal, H.~Kim, V.~Haenel, I.~Fruend, P.~Yianilos, M.~Mueller-Freitag \emph{et~al.}, ``The" something something" video database for learning and evaluating visual common sense,'' in \emph{Proceedings of the IEEE international conference on computer vision}, 2017, pp. 5842--5850.

\bibitem{matterport3d}
A.~Chang, A.~Dai, T.~Funkhouser, M.~Halber, M.~Niessner, M.~Savva, S.~Song, A.~Zeng, and Y.~Zhang, ``Matterport3d: Learning from rgb-d data in indoor environments,'' \emph{arXiv preprint arXiv:1709.06158}, 2017.

\bibitem{nwm}
A.~Bar, G.~Zhou, D.~Tran, T.~Darrell, and Y.~LeCun, ``Navigation world models,'' \emph{arXiv preprint arXiv:2412.03572}, 2024.

\bibitem{SCAND}
H.~Karnan, A.~Nair, X.~Xiao, G.~Warnell, S.~Pirk, A.~Toshev, J.~Hart, J.~Biswas, and P.~Stone, ``Socially compliant navigation dataset (scand): A large-scale dataset of demonstrations for social navigation,'' \emph{IEEE Robotics and Automation Letters}, vol.~7, no.~4, pp. 11\,807--11\,814, 2022.

\bibitem{tartandrive}
S.~Triest, M.~Sivaprakasam, S.~J. Wang, W.~Wang, A.~M. Johnson, and S.~Scherer, ``Tartandrive: A large-scale dataset for learning off-road dynamics models,'' in \emph{2022 International Conference on Robotics and Automation (ICRA)}.\hskip 1em plus 0.5em minus 0.4em\relax IEEE, 2022, pp. 2546--2552.

\bibitem{RECON}
D.~Shah, B.~Eysenbach, G.~Kahn, N.~Rhinehart, and S.~Levine, ``Rapid exploration for open-world navigation with latent goal models,'' \emph{arXiv preprint arXiv:2104.05859}, 2021.

\bibitem{HuRoN}
N.~Hirose, D.~Shah, A.~Sridhar, and S.~Levine, ``Sacson: Scalable autonomous control for social navigation,'' \emph{IEEE Robotics and Automation Letters}, vol.~9, no.~1, pp. 49--56, 2023.

\bibitem{pdsketch}
J.~Mao, T.~Lozano-P{\'e}rez, J.~Tenenbaum, and L.~Kaelbling, ``Pdsketch: Integrated domain programming, learning, and planning,'' \emph{Advances in Neural Information Processing Systems}, vol.~35, pp. 36\,972--36\,984, 2022.

\bibitem{susie}
K.~Black, M.~Nakamoto, P.~Atreya, H.~Walke, C.~Finn, A.~Kumar, and S.~Levine, ``Zero-shot robotic manipulation with pretrained image-editing diffusion models,'' \emph{arXiv preprint arXiv:2310.10639}, 2023.

\bibitem{cliport}
M.~Shridhar, L.~Manuelli, and D.~Fox, ``Cliport: What and where pathways for robotic manipulation,'' in \emph{Conference on robot learning}.\hskip 1em plus 0.5em minus 0.4em\relax PMLR, 2022, pp. 894--906.

\bibitem{kitchenshift}
E.~Xing, A.~Gupta, S.~Powers, and V.~Dean, ``Kitchenshift: Evaluating zero-shot generalization of imitation-based policy learning under domain shifts,'' in \emph{NeurIPS 2021 Workshop on Distribution Shifts: Connecting Methods and Applications}, 2021.

\bibitem{dreamerv3}
D.~Hafner, J.~Pasukonis, J.~Ba, and T.~Lillicrap, ``Mastering diverse domains through world models,'' \emph{arXiv preprint arXiv:2301.04104}, 2023.

\bibitem{minerl}
W.~H. Guss, B.~Houghton, N.~Topin, P.~Wang, C.~Codel, M.~Veloso, and R.~Salakhutdinov, ``Minerl: A large-scale dataset of minecraft demonstrations,'' \emph{arXiv preprint arXiv:1907.13440}, 2019.

\bibitem{minedreamer}
E.~Zhou, Y.~Qin, Z.~Yin, Y.~Huang, R.~Zhang, L.~Sheng, Y.~Qiao, and J.~Shao, ``Minedreamer: Learning to follow instructions via chain-of-imagination for simulated-world control,'' \emph{arXiv preprint arXiv:2403.12037}, 2024.

\bibitem{dosovitskiy2017carla}
A.~Dosovitskiy, G.~Ros, F.~Codevilla, A.~Lopez, and V.~Koltun, ``Carla: An open urban driving simulator,'' in \emph{Conference on robot learning}.\hskip 1em plus 0.5em minus 0.4em\relax PMLR, 2017, pp. 1--16.

\bibitem{rlbench}
S.~James, Z.~Ma, D.~R. Arrojo, and A.~J. Davison, ``Rlbench: The robot learning benchmark \& learning environment,'' \emph{IEEE Robotics and Automation Letters}, vol.~5, no.~2, pp. 3019--3026, 2020.

\bibitem{gibson}
F.~Xia, A.~R. Zamir, Z.~He, A.~Sax, J.~Malik, and S.~Savarese, ``Gibson env: Real-world perception for embodied agents,'' in \emph{Proceedings of the IEEE conference on computer vision and pattern recognition}, 2018, pp. 9068--9079.

\bibitem{metaworld}
T.~Yu, D.~Quillen, Z.~He, R.~Julian, K.~Hausman, C.~Finn, and S.~Levine, ``Meta-world: A benchmark and evaluation for multi-task and meta reinforcement learning,'' in \emph{Conference on robot learning}.\hskip 1em plus 0.5em minus 0.4em\relax PMLR, 2020, pp. 1094--1100.

\bibitem{ithor}
E.~Kolve, R.~Mottaghi, W.~Han, E.~VanderBilt, L.~Weihs, A.~Herrasti, M.~Deitke, K.~Ehsani, D.~Gordon, Y.~Zhu \emph{et~al.}, ``Ai2-thor: An interactive 3d environment for visual ai,'' \emph{arXiv preprint arXiv:1712.05474}, 2017.

\bibitem{umi}
C.~Chi, Z.~Xu, C.~Pan, E.~Cousineau, B.~Burchfiel, S.~Feng, R.~Tedrake, and S.~Song, ``Universal manipulation interface: In-the-wild robot teaching without in-the-wild robots,'' \emph{arXiv preprint arXiv:2402.10329}, 2024.

\bibitem{vima}
Y.~Jiang, A.~Gupta, Z.~Zhang, G.~Wang, Y.~Dou, Y.~Chen, L.~Fei-Fei, A.~Anandkumar, Y.~Zhu, and L.~Fan, ``Vima: General robot manipulation with multimodal prompts,'' \emph{arXiv preprint arXiv:2210.03094}, vol.~2, no.~3, p.~6, 2022.

\bibitem{code-as-policy}
J.~Liang, W.~Huang, F.~Xia, P.~Xu, K.~Hausman, B.~Ichter, P.~Florence, and A.~Zeng, ``Code as policies: Language model programs for embodied control,'' in \emph{2023 IEEE International Conference on Robotics and Automation (ICRA)}.\hskip 1em plus 0.5em minus 0.4em\relax IEEE, 2023, pp. 9493--9500.

\bibitem{voxposer}
W.~Huang, C.~Wang, R.~Zhang, Y.~Li, J.~Wu, and L.~Fei-Fei, ``Voxposer: Composable 3d value maps for robotic manipulation with language models,'' \emph{arXiv preprint arXiv:2307.05973}, 2023.

\bibitem{rekep}
W.~Huang, C.~Wang, Y.~Li, R.~Zhang, and L.~Fei-Fei, ``Rekep: Spatio-temporal reasoning of relational keypoint constraints for robotic manipulation,'' \emph{arXiv preprint arXiv:2409.01652}, 2024.

\bibitem{gs}
B.~Kerbl, G.~Kopanas, T.~Leimk{\"u}hler, and G.~Drettakis, ``3d gaussian splatting for real-time radiance field rendering.'' \emph{ACM Trans. Graph.}, vol.~42, no.~4, pp. 139--1, 2023.

\bibitem{opendv2k}
J.~Yang, S.~Gao, Y.~Qiu, L.~Chen, T.~Li, B.~Dai, K.~Chitta, P.~Wu, J.~Zeng, P.~Luo \emph{et~al.}, ``Generalized predictive model for autonomous driving,'' in \emph{Proceedings of the IEEE/CVF Conference on Computer Vision and Pattern Recognition}, 2024, pp. 14\,662--14\,672.

\end{thebibliography}
